\documentclass[journal]{IEEEtran}

\IEEEoverridecommandlockouts
\usepackage{cite}
\usepackage{amsmath,amssymb,amsfonts}
\usepackage{algorithmic}
\usepackage{graphicx}
\usepackage{subcaption}
\usepackage{textcomp}
\usepackage{xcolor}
\usepackage{booktabs}
\usepackage{arydshln}
\usepackage{makecell}
\usepackage{pifont}
\usepackage{arydshln}
\newcommand{\cmark}{\ding{51}} 
\newcommand{\xmark}{\ding{55}} 
\def\BibTeX{{\rm B\kern-.05em{\sc i\kern-.025em b}\kern-.08em
    T\kern-.1667em\lower.7ex\hbox{E}\kern-.125emX}}

\begin{document}

\title{Multi-Station WiFi CSI Sensing Framework Robust to Station-wise Feature Missingness and Limited Labeled Data\\
  \thanks{Keita Kayano and Takayuki Nishio are with the School of Engineering, Institute of Science Tokyo, 2-12-1 Ookayama, Tokyo, 152-8550, Japan (e-mail: nishio@ict.eng.isct.ac.jp). Daiki Yoda, Yuta Hirai, and Tomoko Adachi are with Wireless Systems R\&D Department, Infrastructure Systems R\&D Center, Corporate Laboratory, Toshiba Corporation, 1, Komukai Toshiba-cho, Saiwai-ku, Kawasaki 212-8582, JAPAN. This work was supported in part by JSPS KAKENHI Grant Numbers JP23K24831 and JP23K26109.
  This work has been submitted to the IEEE for possible publication. Copyright may be transferred without notice, after which this version may no longer be accessible.}
}

\author{Keita~Kayano,
        Takayuki~Nishio,~\IEEEmembership{Senior~Member,~IEEE,}\\
        Daiki~Yoda,~\IEEEmembership{Member,~IEEE,}
        Yuta~Hirai,
        and~Tomoko~Adachi.}

\maketitle
\begin{abstract}
We propose a WiFi Channel State Information (CSI) sensing framework for multi-station deployments that addresses two fundamental challenges in practical CSI sensing: station-wise feature missingness and limited labeled data.
Feature missingness is commonly handled by resampling unevenly spaced CSI measurements or by reconstructing missing samples, while label scarcity is mitigated by data augmentation or self-supervised representation learning. 
However, these techniques are typically developed in isolation and do not jointly address long-term, structured station unavailability together with label scarcity. 
To bridge this gap, we explicitly incorporate station unavailability into both representation learning and downstream model training.
Specifically, we adapt cross-modal self-supervised learning (CroSSL), a representation learning framework originally designed for time-series sensory data, to multi-station CSI sensing in order to learn representations that are inherently invariant to station-wise feature missingness from unlabeled data.
Furthermore, we introduce Station-wise Masking Augmentation (SMA) during downstream model training, which exposes the model to realistic station unavailability patterns under limited labeled data.
Our experiments show that neither missingness-invariant pre-training nor station-wise augmentation alone is sufficient; their combination is essential to achieve robust performance under both station-wise feature missingness and label scarcity.
The proposed framework provides a practical and robust foundation for multi-station WiFi CSI sensing in real-world deployments.

\end{abstract}


\section{Introduction}

WiFi sensing has emerged as a promising environmental sensing technology that leverages variations in WiFi radio signals caused by human motion and object interactions in the propagation environment \cite{csi_sensing_survey, csi_sensing_survey2}.
By exploiting existing WiFi infrastructure, WiFi sensing enables non-intrusive sensing at low deployment cost and is robust to lighting conditions.
Among various approaches, Channel State Information (CSI) sensing has demonstrated particularly strong performance due to its ability to capture fine-grained channel variations beyond conventional Received Signal Strength Indicator based methods.
Combined with machine learning (ML), CSI sensing has enabled a wide range of applications, including human localization, pose estimation, and image generation \cite{localization, multi_station_pose_estimation, GenImage}.

Despite these advances, practical CSI sensing faces two major challenges: feature missingness and limited labeled data.
First, CSI sensing systems often suffer from missing features, particularly in multi-station deployments.
Most prior CSI sensing systems rely on a single station \cite{MMFi, dataset:WiMANS}, which simplifies deployment and learning.
To further improve sensing performance, recent studies have started to exploit multi-station CSI, aggregating measurements from spatially distributed stations to capture complementary viewpoints \cite{multi_station_pose_estimation, WiPose}.
In many of these settings, the input is implicitly assumed to be fully observed and temporally aligned across stations.
In practice, however, some stations' CSI are unavailable for extended periods due to sparse and heterogeneous, application-driven transmissions and network contention, resulting in station-wise feature missingness.
This structured missingness induces a distribution shift between training and inference, causing severe performance degradation for models trained under complete-input assumptions.
To cope with feature missingness, prior studies have proposed resampling-based approaches that construct uniformly sampled CSI sequences from unevenly spaced measurements \cite{linear_interpolation1, nearest_interpolation2}. 
In addition, reconstruction-based methods have been explored to restore missing CSI samples before feeding them into downstream models \cite{recover_multiCSI}.
These techniques are effective for compensating short-term missingness and irregular sampling.

Second, labeled data are often scarce in CSI sensing.
CSI characteristics are highly environment-dependent, and models trained in one environment do not readily generalize to new settings.
Consequently, reliable performance typically requires environment-specific labeled data.
However, collecting large-scale labeled CSI datasets is costly and labor-intensive, making fully supervised learning impractical in many scenarios.
To mitigate label dependency, prior studies have explored data augmentation, which synthetically expands the effective training set by perturbing CSI samples while preserving task-relevant information\cite{aug_time1, aug_time2, aug3, aug4, aug5}.
More recently, Self-Supervised Learning (SSL) has gained attention as a complementary approach that learns task-agnostic representations from unlabeled CSI and then trains a downstream model with limited labeled data \cite{csi_ssl_survey}.
Recent SSL-based CSI sensing methods have demonstrated promising label efficiency \cite{autofi, multi_device_ssl, CSI_SSL_HAR}.

However, these two challenges have largely been addressed in isolation.
Methods for handling missing CSI typically assume sufficient labeled data for downstream training, while label-efficient approaches are commonly developed under complete-input assumptions.
As a result, existing methods struggle when station-wise feature missingness and label scarcity occur simultaneously in real deployments.

To bridge this gap, we propose a CSI sensing framework that jointly addresses station-wise feature missingness and limited labeled data.
The key idea of our approach is to learn representations that are inherently invariant to station-wise missingness while simultaneously reducing reliance on labeled data.
This is achieved by explicitly enforcing consistency across different station-availability patterns during self-supervised pre-training.
We argue that learning representations invariant to station availability during SSL pre-training not only mitigates feature missingness but also yields task-agnostic features that remain effective under limited labeled data.
To this end, we employ cross-modal self-supervised learning (CroSSL) \cite{crossl}, a representation learning framework originally designed for time-series sensor data for jointly mitigating station-wise feature missingness and limited labeled data.
By explicitly learning representations of CSI under different station-missing patterns, the pre-trained feature extractor becomes inherently robust to station-wise feature missingness.
Furthermore, we extend this principle to downstream model training by introducing Station-wise Masking Augmentation (SMA).
SMA applies the same missingness-aware strategy to labeled data, ensuring that the downstream model is exposed to realistic station unavailability patterns during training.

We validate the proposed framework using two real-world multi-station CSI datasets collected by ourselves in an office-like environment and a factory-like environment.
In both settings, multi-station CSI measurements were acquired using commodity devices compliant with IEEE~802.11 rather than relying on simulations.
Extensive experiments demonstrate that the proposed method consistently outperforms existing pre-training and data augmentation baselines under both station-wise feature missingness and limited labeled data. 
The results further indicate that neither missingness-invariant pre-training nor station-wise augmentation alone is sufficient; rather, their combination is essential to achieve robust performance under these conditions.

Our contributions are summarized as follows.
\begin{enumerate}
    \item We identify the gap between existing CSI sensing frameworks and practical multi-station deployments, highlighting station-wise feature missingness and label scarcity as key challenges.
    \item We propose a unified CSI sensing framework that combines missingness-invariant self-supervised pre-training with station-aware data augmentation.
    \item We demonstrate the effectiveness of the proposed method across different environments and downstream tasks through comprehensive experimental evaluation.
    These experiments are conducted on real-world multi-station CSI datasets collected using commodity devices.
\end{enumerate}


\section{Related Work}

\begin{table*}[t]
\centering
\caption{Comparison of related CSI sensing methods in terms of missingness handling and learning assumptions.}
\label{tab:related_work_summary}
\begin{tabular}{lccccc}
\toprule
Method 
& Approaches
& \#Stations
& \makecell{Handles Station-wise \\ Missingness}
& \makecell{Works with \\ Uncontrolled Transmissions}
& Label-efficient 
\\
\midrule
RFboost~\cite{csi_randomerasing} 
& DA
& -
& - 
& -
& \cmark
\\

RR-scheduled CSI recovery with inpainting~\cite{recover_multiCSI} 
& DA
& Multi
& \cmark 
& \xmark
& \xmark
\\

CSI-BERT~\cite{CSI_BERT} 
&  DA
&  Single
&  \xmark
&  \cmark
&  \xmark
\\

AutoFi~\cite{autofi} 
& Pre-training
& Single
& \xmark
& \xmark 
& \cmark
\\

Multi-device SSL~\cite{multi_device_ssl} 
& Pre-training
& Multi
& \xmark
& \xmark 
& \cmark
\\

\midrule
\textbf{Ours}
& Pre-training + DA
& Multi
& \cmark
& \cmark
& \cmark \\
\bottomrule
\end{tabular}
\end{table*}

In this section, we summarize related studies on WiFi CSI sensing. Table \ref{tab:related_work_summary} compares the key related works and highlights the differences between them and the proposed method. Details are discussed in the following subsections.

\subsection{Handling Feature Missingness in CSI Sensing}
Handling feature missingness is a fundamental challenge in practical CSI sensing. 
A common and widely studied problem is the handling of unevenly sampled frames at the packet level. 
In practical deployments, even if a transmitter is programmed to send packets at a fixed interval, the receiver may not capture them at the same frequency. 
This sampling jitter is common and can be caused by various factors, including packet loss or non-real-time operating systems on the access point (AP) that cannot guarantee task priority \cite{csi_jitter}. 
These jitters can sometimes result in gaps of up to several hundred milliseconds.

To create a uniform time series required by most ML models, data interpolation is widely employed as a resampling technique. 
Methods such as Linear Interpolation \cite{linear_interpolation1, linear_interpolation2, linear_interpolation3} and Nearest Neighbor Interpolation \cite{WiPose, nearest_interpolation2} are commonly used to estimate the CSI values at a evenly spaced sequence of timestamps.
More advanced, deep learning-based approaches have also been proposed \cite{CSI_BERT, Gan-based_upsampling}. 
Zhao et al. introduced CSI-BERT, which adapts Bidirectional Encoder Representations from Transformer to CSI imputation by directly projecting continuous CSI measurements into token embeddings using a linear layer, thereby avoiding information loss caused by discretization \cite{CSI_BERT}.

However, these methods are fundamentally designed to address short-term packet-level irregularities.
Their scope is limited to resampling and correcting local temporal gaps.
Applying interpolation or imputation across long-term station-wise missingness, which spans several seconds or more, would introduce unrealistic artifacts and distort the underlying signal structure.

More recently, Pyo and Choi proposed a multi-link CSI sensing framework to mitigate CSI loss in congested network environments~\cite{recover_multiCSI}.
Their approach explicitly controls CSI acquisition at the AP by applying Round-Robin scheduling across multiple links, thereby coordinating which station transmits CSI at each time slot and reducing network contention.
To further compensate for CSI samples lost due to packet drops, a Context Encoder–based deep inpainting model \cite{contextencoder} is employed to reconstruct missing CSI before downstream processing.

Overall, prior work on feature missingness has largely focused on resampling or reconstructing CSI to recover a complete input stream.
However, these pipelines do not directly address the complementary challenge of limited labeled data, since the downstream model still requires sufficient supervision after resampling or recovery.
This gap motivates our approach of learning station-missingness-invariant representations from unlabeled CSI, enabling robust sensing under both station-wise missingness and label scarcity.

\subsection{Data Augmentation in CSI Sensing}
Data augmentation (DA) is a widely used technique to artificially increase the amount of training data by applying various transformations to the original samples.
In the context of CSI sensing, temporal domain transformations have been particularly effective for improving model generalization.
Typical approaches include shifting the observation window along the time axis and applying time warping to stretch or compress the temporal scale of the signal\cite{aug_time1, aug_time2, aug3, aug4, aug5}.
Such transformations help models adapt to variations in motion speed and activity patterns.
In addition to temporal transformations, other simple augmentation methods such as amplitude scaling, Gaussian noise injection, and frequency-domain shifting have also been explored.
These methods aim to enhance the diversity of training data and improve the robustness of the learned representations to non-informative fluctuations in CSI that do not reflect the underlying sensing target.

Whereas previous methods distort or add noise to the entire signal, another line of research improves robustness by removing parts of the input. Inspired by techniques in computer vision (CV) that mask input regions \cite{cutout, RandomErasing} to prevent models from overly relying on specific local features, this concept has also been applied to CSI sensing.
Hou et al. proposed Motion-aware Random Erasing (MRE) \cite{csi_randomerasing}, which adapts object-aware erasing strategies from CV to the CSI domain. Unlike simple random time/frequency masking, MRE places erasure masks specifically within motion-rich regions detected in the spectrogram. This encourages the model to focus on more overall features rather than overfitting to localized, motion-specific patterns, thereby improving generalization by deliberately obscuring details that may not generalize.

Beyond modifying existing samples, DA approaches based on deep generative models have attracted increasing attention\cite{cgan, generative_csiaug}.
Wang et al. proposed a diffusion model–based DA framework for enhancing spectra data quality\cite{generative_csiaug}.
In their approach, a conditional diffusion model is first trained on real-world samples to generate acceleration-jerk spectra derived from CSI measurements.
Subsequently, another diffusion model is trained using these generated spectra to reduce noise and improve fidelity.
This two-stage training process enables the generation of high-quality spectra from a limited amount of real data, thereby improving data diversity and quality for downstream tasks.

In contrast, our proposed approach introduces a simple yet effective augmentation strategy specifically designed to enhance robustness against station-wise feature missingness. 
By simulating station-wise feature missingness scenarios during training, our method enables the model to remain stable even under incomplete input conditions.

\subsection{SSL in CSI Sensing}
SSL is a branch of unsupervised learning that learns useful representations through pretext tasks using unlabeled samples.
By pre-training a feature extractor with SSL, the reliance on a large number of labeled samples for downstream tasks can be greatly reduced.
In addition, since SSL learns task-agnostic representations, it is also effective for domain adaptation to different environments and scenarios.

Several studies have explored the use of SSL for CSI sensing.
Yang et al. proposed AutoFi\cite{autofi}, an SSL-based CSI sensing model that enables efficient adaptation to new environments.
AutoFi learns consistency between embeddings obtained from two views generated by adding Gaussian noise to the input data.
Its loss function is a weighted sum of three components: probability consistency loss, mutual information loss and geometric consistency loss.
In particular, the geometric consistency loss preserves the geometric structure between samples in the embedding space, improving domain adaptation performance with a limited amount of labeled data.
Bocus et al. proposed a model that treats CSI signals acquired simultaneously from two receivers as different views and learns consistency between their representations\cite{multi_device_ssl}.
This approach applies the principles of SimCLR\cite{SimCLR} and Contrastive Multiview Coding \cite{contrastivemultiviewcoding} to CSI sensing.

While these approaches enable representation learning from unlabeled samples, they implicitly assume that all stations are synchronized and that no feature missingness occurs.
As a result, they are vulnerable to station-wise feature missingness.
In contrast, our proposed method leverages the advantages of SSL while learning station-missingness-invariant representations.

\section{Proposed Method}

\begin{figure}
    \centering
    \includegraphics[width=0.9\linewidth]{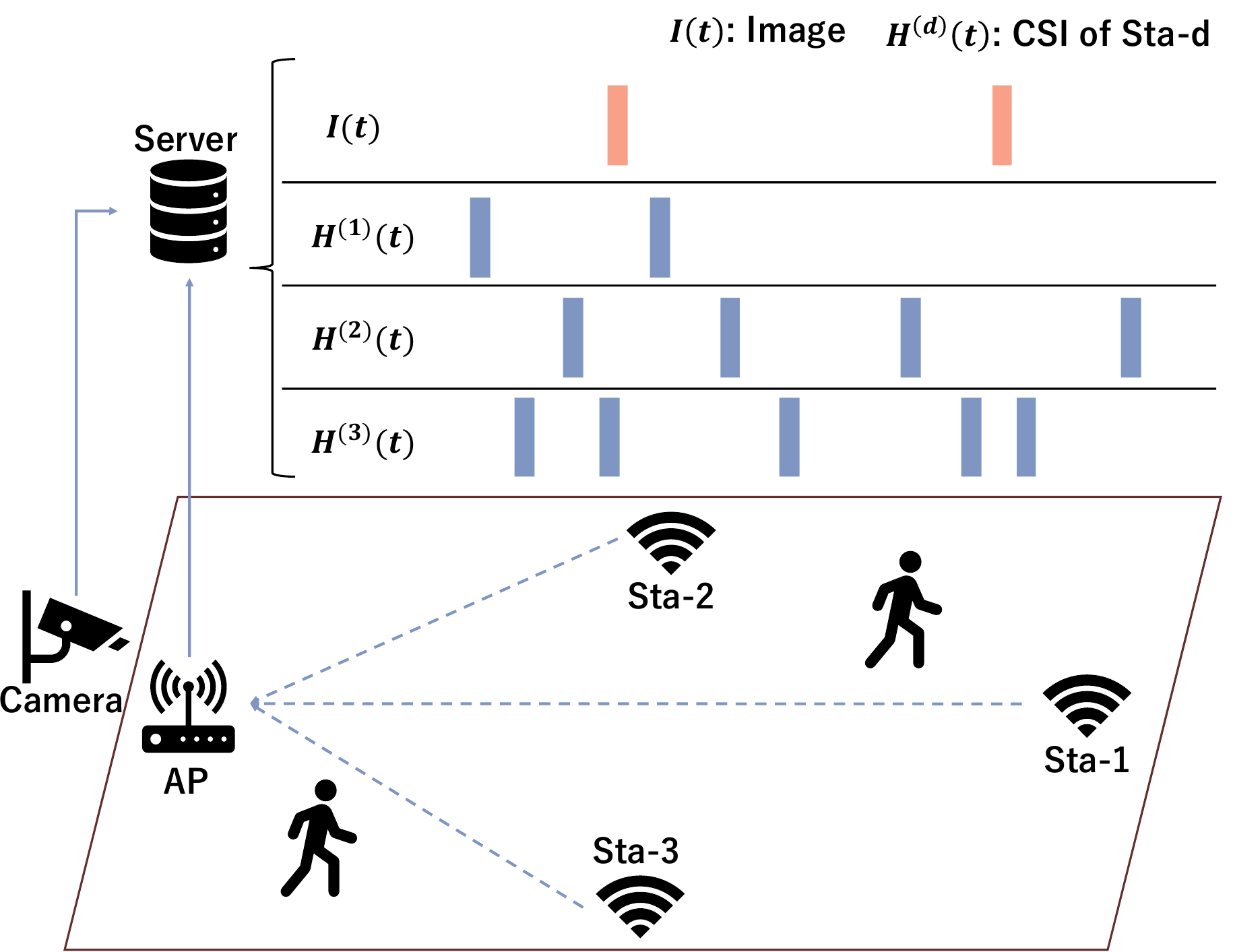}
    \caption{Multi-stations CSI sensing setup and visualization of timestamp alignment. 
    Multiple stations transmit frames to the AP. 
    A camera provides images for ground-truth of downstream model training. 
    As illustrated in the upper timeline, WiFi frames from different stations and camera images are acquired asynchronously and at heterogeneous intervals.
    }
    \label{fig:collect_data}
\end{figure}

\begin{figure}
    \centering
    \includegraphics[width=0.9\linewidth]{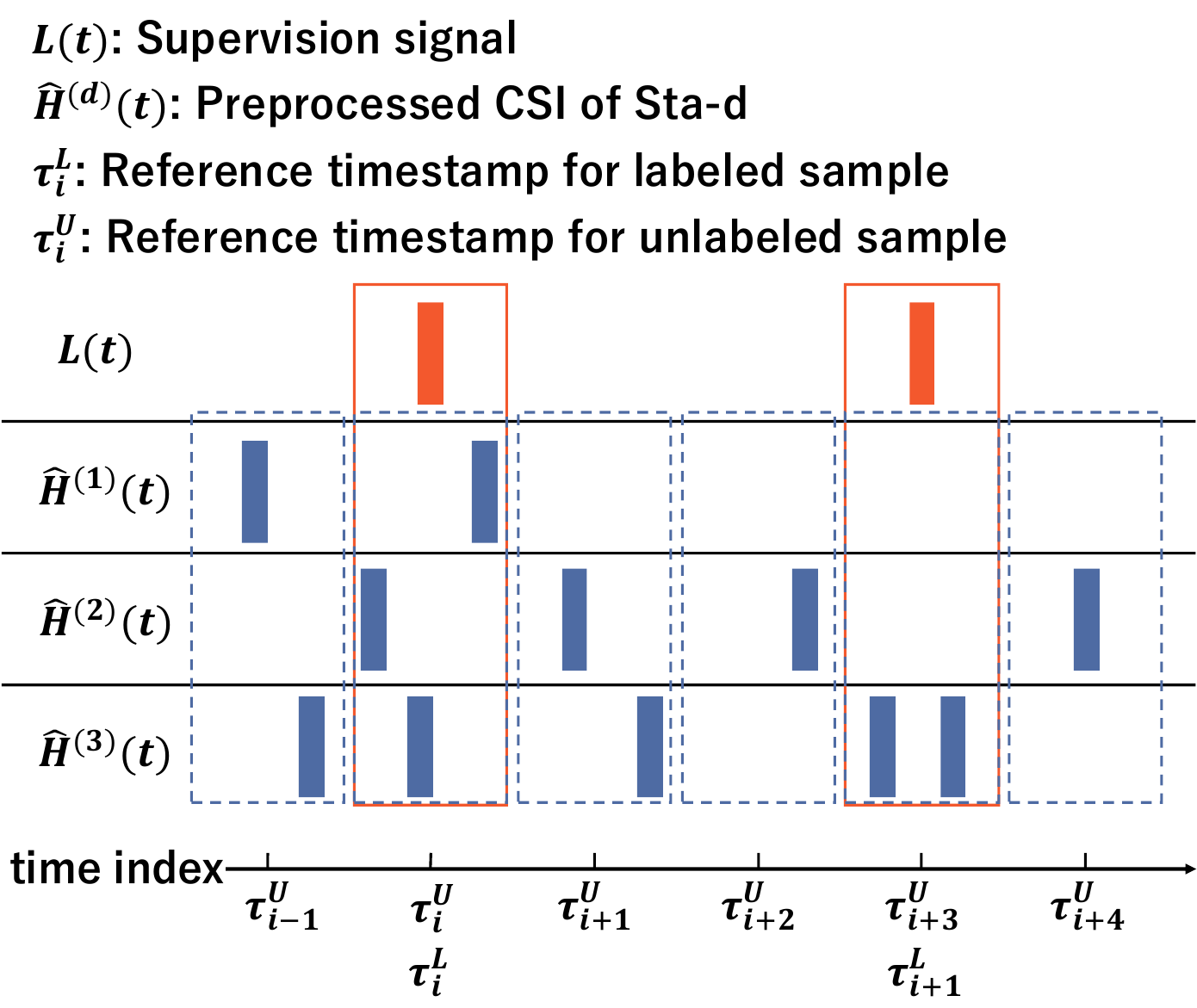}
    \caption{Illustration of labeled and unlabeled sample generation from asynchronous CSI and label streams.
    Both labeled and unlabeled samples are constructed by applying a fixed-length window around reference timestamps.
    Due to asynchronous sensing and heterogeneous transmission intervals, some windows may contain no CSI from particular stations, resulting in station-wise feature missingness.}
    \label{fig:generate_sample}
\end{figure}

\subsection{Assumption}\label{subsec:assumption}
In this study, we consider a practical multi-station CSI sensing scenario in which CSI from some stations may be unavailable at inference time and labeled data are limited.
Fig.~\ref{fig:collect_data} shows the assumed multi-station WiFi CSI sensing setup.
Multiple fixed stations transmit WiFi frames to a single AP which estimates CSI from received signals.
A camera, deployed independently of the wireless system, captures images $I(t)$ at time $t$.
The images are used to derive ground-truth for downstream tasks.
CSI acquisition and image capture are asynchronous.

We next formalize the wireless signal model and define the CSI used throughout this study.
The relationship between the transmitted signal $x^{(d)}(t,k)$ and the received signal $y^{(d)}(t,k)$ of station $d$ at time $t$ and subcarrier $k$ is expressed as
\begin{equation}
y^{(d)}(t,k) = h^{(d)}(t,k)x^{(d)}(t,k) + n^{(d)}(t,k),
\label{eq:csi_difinition}
\end{equation}
where $h^{(d)}(t,k)$ denotes the channel frequency response and $n^{(d)}(t,k)$ represents additive noise.
The matrix $H^{(d)}(t)$, which concatenates $h^{(d)}(t,k)$ over all subcarriers, is referred to as the Channel State Information (CSI).
CSI can be estimated from known shared pilot symbols.
Since CSI is a complex-valued matrix, it can be decomposed into amplitude and phase components as
\begin{equation}
H^{(d)}(t) = \|H^{(d)}(t)\| e^{j\angle H^{(d)}(t)},
\label{eq:csi_complex}
\end{equation}
where $\|H^{(d)}(t)\|$ and $\angle H^{(d)}(t)$ denote the amplitude and phase of $H^{(d)}(t)$, respectively.
Since variations in the propagation environment are reflected in the amplitude attenuation and phase shift of CSI, it can be applied to a wide range of sensing tasks.
However, in commercial WiFi devices, accurately obtaining phase information is challenging due to Sampling Frequency Offset and other hardware impairments\cite{CSI_Phase}.
Therefore, we use only the amplitude in this study.

The collected CSI and camera images are transmitted to a server.
To transform these raw measurements into stable inputs suitable for learning-based sensing,
standard preprocessing and sampling procedures are applied, as described below.
CSI preprocessing consists of subcarrier selection and scaling.
Specifically, we remove null subcarriers and the DC subcarrier to eliminate irrelevant components \cite{remove_subcarrier}.
For scaling, we apply a power normalization method based on the average power of each sample, which is defined as follows:
\begin{equation}
    \widehat{h}^{(d)}(t,k) = \frac{\| h^{(d)}(t,k) \|}{[ {\frac{1}{K}\sum_{1\le k \le K}\| h^{(d)}(t,k) \|^2} ] ^{1/2}},
\end{equation}
where $K$ denotes the number of selected subcarriers.
The matrix $\widehat{H}^{(d)}(t)$, which concatenates $\widehat{h}^{(d)}(t,k)$ over all subcarriers, is referred to as the preprocessed CSI.
This method assumes that variations in amplitude scale primarily originate from gain errors.
The stability and effectiveness of this normalization preprocessing were demonstrated in \cite{csi_preprocess} through simulation studies and measurement experiments spanning several hours.

Image preprocessing depends on the downstream task.
A task-specific labeling function
$g(\cdot)$ is applied to the image stream to generate a label signal
\begin{equation}
    L(t) = g\!\left(I(t)\right),
\end{equation}
where $L(t)$ represents a continuous-time supervision signal.
For example, $L(t)$ may correspond to the image itself in image generation tasks,
or to extracted quantities such as human positions in localization tasks.


After preprocessing, both labeled and unlabeled samples are generated, as illustrated in Fig.~\ref{fig:generate_sample}.
Let $\{\tau_i^{\mathrm{L}}\}$ denote the reference timestamps determined by the label acquisition rate $R_{\mathrm{Label}}$.
For each reference timestamp $\tau_i^{\mathrm{L}}$, the corresponding ground-truth $Y_i$ is obtained
by associating the temporally nearest $L(t)$.

For CSI, a fixed-length temporal window of width $w$ centered at $\tau_i^{\mathrm{L}}$ is applied.
Let $\mathcal{W}_i = [\tau_i^{\mathrm{L}} - \tfrac{w}{2}, \, \tau_i^{\mathrm{L}} + \tfrac{w}{2}]$ denote this window.
For station $d$, CSI frames whose timestamps fall within $\mathcal{W}_i$ are aggregated by temporal averaging:
\begin{equation}
X_i^{(d)} = 
\begin{cases}
\displaystyle
\frac{1}{|\mathcal{S}_i^{(d)}|}
\sum\limits_{\widehat{H}^{(d)}(t)\in\mathcal{S}_i^{(d)}} \widehat{H}^{(d)}(t),
& \text{if } |\mathcal{S}_i^{(d)}| > 0, \\
M_{inp}, & \text{otherwise},
\end{cases}
\end{equation}
where $\mathcal{S}_i^{(d)}$ denotes the set of CSI samples from station $d$ within $\mathcal{W}_i$,
and $M_{inp}$ denotes a masked value indicating the absence of CSI from that station.
In this study, we use zero-padding as $M_{inp}$.
This window-based temporal aggregation is widely used in CSI sensing
to extract stable statistical features from noisy packet-level measurements\cite{csi_sensing_survey2}.
The multi-station CSI input for sample $i$ is constructed by stacking all stations:
\begin{equation}
X_i = \big[ X_i^{(1)}, X_i^{(2)}, \ldots, X_i^{(N_d)} \big].
\end{equation}
The labeled sample $i$ is thus defined as a pair of $(X_i, Y_i)$.
This formulation leads to a characteristic form of missing data
in multi-station CSI sensing.

For a given sample $i$, we define the set of missing stations as
\begin{equation}
\mathcal{M}_i^{\mathrm{obs}} = \left\{ d \in \{1,\dots,N_d\} \;\middle|\; |\mathcal{S}_i^{(d)}| = 0 \right\}.
\end{equation}
A station $d$ is said to be missing if no CSI from that station
is observed within the aggregation window $\mathcal{W}_i$.
Station-wise feature missingness is said to occur for sample $i$
when $\mathcal{M}_i^{\mathrm{obs}}$ is non-empty, i.e., $|\mathcal{M}_i^{\mathrm{obs}}| > 0$.

Station-wise feature missingness naturally arises in practical
multi-station CSI sensing systems because CSI acquisition depends
on application-driven transmissions from each station.
Unlike centralized sensing setups, stations transmit frames
asynchronously and opportunistically for their own communication
purposes, and the sensing system has no control over their timing.
As a result, it is common that no CSI frame from a given station
is observed within a temporal aggregation window.

Unlabeled samples are generated in a similar manner but without relying on label timestamps.
Instead, reference timestamps $\{\tau_i^{\mathrm{U}}\}$ are uniformly sampled at a rate $R_{\mathrm{SSL}}$,
where $R_{\mathrm{SSL}} > R_{\mathrm{Label}}$.
For each $\tau_i^{\mathrm{U}}$, CSI aggregation is performed using the same windowing and averaging procedure as above.
These preprocessing and sampling procedures are widely adopted in CSI sensing literature and are treated as common assumptions rather than task-specific designs.

\begin{figure*}[]
    \centering
    \includegraphics[width=0.95\linewidth]{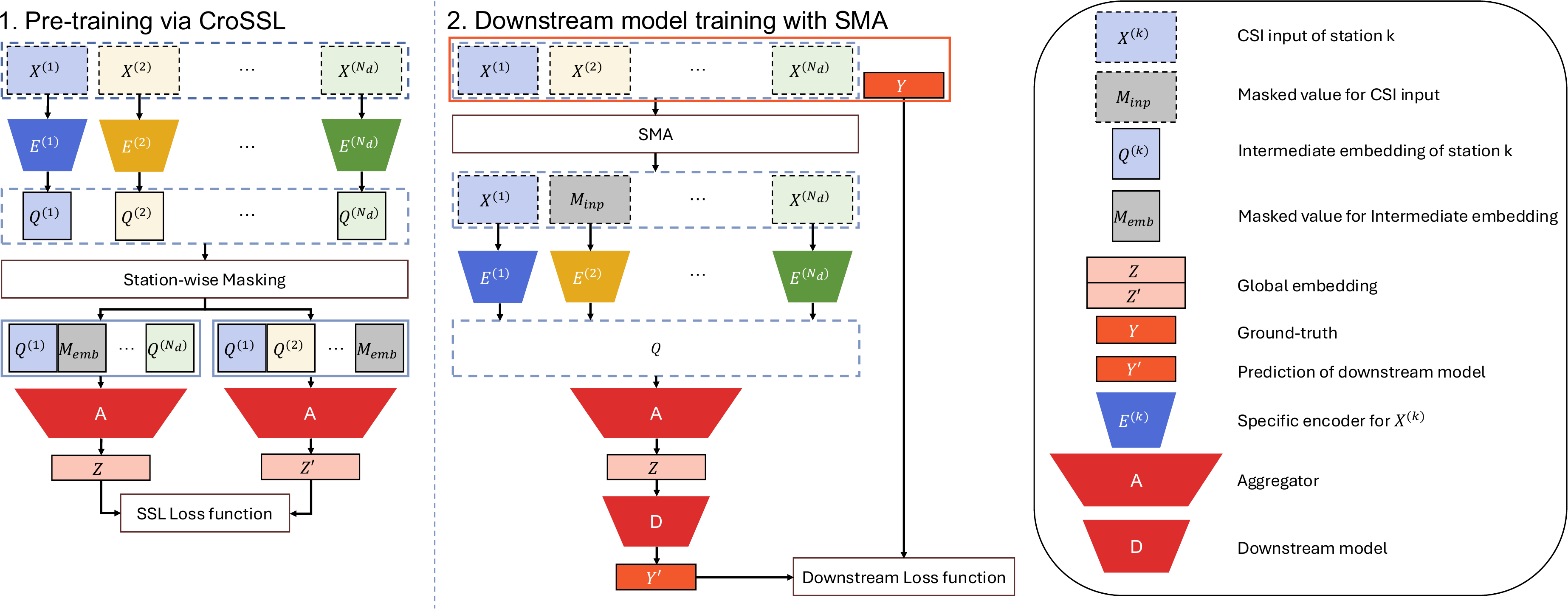}
    \caption{
    Overview of the proposed framework.
    The framework learns station-missingness-invariant representations from multi-station CSI and applies them to downstream tasks.
    CroSSL is used to learn robust global embeddings from unlabeled CSI by  masking intermediate station embeddings during representation learning.
    For downstream model training, SMA  masks entire station inputs to simulate realistic station unavailability.
    The same encoders and aggregator are shared across both phases, enabling robust inference under station-wise feature missingness and limited labeled data.}
    \label{fig:framework}
\end{figure*}

\subsection{Framework Overview}

To address both station-wise feature missingness and limited labeled data in multi-station CSI sensing, we propose a unified learning framework, illustrated in Fig.~\ref{fig:framework}.
All variables in Fig.~\ref{fig:framework} are defined at the sample level; the subscript $i$ is omitted in the figure for readability.

The proposed framework consists of two steps: pre-training with an unlabeled dataset and downstream model training with a labeled dataset.
Here, we briefly describe the training procedure, with details deferred to SubSection \ref{subsec:pre-train}, \ref{subsec:downstream}. 
In the pre-training, we aim to learn a station-missingness-invariant global representation from unlabeled CSI data.
To explicitly model station-wise missingness during pre-training, we employ cross-modal self-supervised learning (CroSSL)~\cite{crossl}.
Each station-specific input $X_i^{(d)}$ is encoded by $E^{(d)}$ to obtain an intermediate embedding ${Q_i^{(d)}}$. 
To simulate station-wise feature missingness, two artificial masking sets
$\mathcal{M}_i^{\mathrm{mask}},\mathcal{M}_i^{\mathrm{mask}'}  \subseteq \{1,\ldots,N_d\}$ are sampled,
and the embeddings corresponding to stations in $\mathcal{M}_i^{\mathrm{mask}}, \mathcal{M}_i^{\mathrm{mask}'}$ are replaced with a masked value $M_{\mathrm{emb}}$.
The masked embeddings are then fused by a shared aggregator $A(\cdot)$ to produce two global embeddings, $Z_i$ and $Z_i'$. 
A self-supervised loss is imposed between $Z_i$ and $Z_i'$, encouraging consistent global representations under different station-wise missingness patterns. 
By enforcing agreement between representations derived from different subsets of available stations, the model is discouraged from relying on any specific station and instead learns station-missingness-invariant global representations.
As a result, the feature extractor consisting of $E^{(d)}(\cdot)$ and $A(\cdot)$ learns representations invariant to station-wise feature missingness while leveraging unlabeled data.

Next, the downstream model is trained with limited labeled data.
While CroSSL encourages the feature extractor to produce representations invariant to missing stations, a mismatch can still arise if the downstream model is trained only on fully observed multi-station inputs.
To address this issue, we introduce Station-wise Masking Augmentation (SMA) during downstream model training.
Given a labeled multi-station input $X_i$ with ground-truth $Y_i$,
we first sample a station-wise masking set
$\mathcal{M}_i^{\mathrm{mask}} \subseteq \{1,\ldots,N_d\}$.
The inputs corresponding to stations in $\mathcal{M}_i^{\mathrm{mask}}$
are then replaced with a masked value $M_{\mathrm{inp}}$.
The masked input is processed by the same station-specific encoders and aggregator as in pre-training to obtain a global embedding $Z_i$,
which is then fed into a downstream model $D(\cdot)$ to produce a prediction $\hat{Y}_i$.
The downstream model is trained by minimizing a supervised loss between
$\hat{Y}_i$ and the ground-truth $Y_i$.
By explicitly exposing the downstream model to station-wise missingness during training, SMA ensures that the entire sensing model remains robust when stations are unavailable at inference time.


The key insight of the proposed framework is that both stages are essential and complementary.
CroSSL enables learning missingness-invariant representations under limited labeled data by leveraging unlabeled CSI, while SMA propagates this robustness to the downstream model by aligning the training and inference conditions.
As demonstrated in our experiments, omitting either component leads to degraded performance, highlighting the necessity of their joint design.

\subsection{Pre-training via CroSSL}\label{subsec:pre-train}
The pre-training stage aims to learn a feature extractor that yields station-missingness-invariant representations from multi-station CSI by utilizing CroSSL-based training framework \cite{crossl}. Since CroSSL was originally designed for time-series sensory data such as accelerometers, gyroscopes, and biosignals (e.g., heart rate, electroencephalograms, electromyograms, electrooculograms, and electrodermal signals), we adapt it for our target application of WiFi CSI sensing.

Each station input is processed by a station-specific encoder $E^{(d)}(\cdot)$, yielding an intermediate embedding
\begin{equation}
Q_i^{(d)} = E^{(d)}\bigl(X_i^{(d)}\bigr).
\end{equation}
The set of intermediate embeddings is denoted as ${Q}_i = [ Q_i^{(1)}, Q_i^{(2)}, \ldots, Q_i^{(N_d)} ]$.
To simulate station-wise feature missingness during pre-training,
we apply station-wise masking to the intermediate embeddings.
Specifically, for each view, a subset of stations is randomly selected
and their corresponding embeddings are replaced with a masked value $M_{\mathrm{emb}}$.

Let $\mathcal{M}_i^{\mathrm{mask}} \subseteq \{1, \ldots, N_d\}$ denote the set of masked stations.
The masked embedding $\widetilde{Q}_i^{(d)}$ is defined as
\begin{equation}
\widetilde{Q}_i^{(d)} =
\begin{cases}
Q_i^{(d)}, & d \notin \mathcal{M}_i^{\mathrm{mask}}, \\
M_{\mathrm{emb}}, & d \in \mathcal{M}_i^{\mathrm{mask}}.
\end{cases}
\end{equation}
In this study, we use zero-padding as $M_{emb}$.
The masking set $\mathcal{M}_i^{\mathrm{mask}}$ is determined by a masking probability $p_{\mathrm{mask}}$.
Specifically, for each station $d$, its embedding is independently masked with probability
$p_{\mathrm{mask}}$, i.e.,
\begin{equation}
\mathbb{P}(d \in \mathcal{M}_i^{\mathrm{mask}}) = p_{\mathrm{mask}}.
\end{equation}

Two masked views are independently generated by applying different masking patterns,
resulting in $\widetilde{Q}_i$ and $\widetilde{Q}_i'$.
Each view is then aggregated by a shared aggregator $A(\cdot)$ to produce
global embedding:
\begin{equation}
Z_i = A(\widetilde{Q}_i), \quad
Z_i' = A(\widetilde{Q}_i').
\end{equation}
By encouraging $Z_i$ and $Z_i'$ to be close to each other while preventing representation collapse (a failure mode where the model outputs the same trivial embedding for all inputs), the model learns station-missingness-invariant embeddings.
To achieve this, an appropriate loss function is required to align the embeddings while avoiding trivial solutions.

In self-supervised learning, loss functions can be broadly categorized into contrastive and non-contrastive methods.
Contrastive methods bring similar samples (positive pairs) closer and push dissimilar samples (negative pairs) apart to prevent representation collapse.
However, the effectiveness of these methods heavily depends on the quality of positive and negative pairs.
In the case of CSI, the low interpretability of the signal makes it difficult to construct reliable positive and negative pairs, and false labeling can significantly degrade performance.
In contrast, non-contrastive methods avoid using explicit negative samples and prevent collapse through model architectural asymmetry or explicit regularization in the embedding space.
In this study, we adopt a variant of VICReg\cite{vicreg}, which is also employed in CroSSL\cite{crossl}, as a non-contrastive method to learn stable and diverse representations.
VICReg consists of three components: variance, invariance, and covariance regularization terms, which together promote both representation stability and information diversity in self-supervised learning.
The variance regularization term is defined as
\begin{equation}
    v(Z) = \frac{1}{l} \sum_{j=1}^{l} {\mathrm{max}(0, \gamma - S(z_{:,j}, \epsilon))},
\end{equation}
where $l$ is the embedding dimension and $z_{:,j}$ denotes the vector of the j-th feature across a batch of size $n$.
The constant $\gamma$ represents the target standard deviation, which is fixed at 1 in our experiments.
$S(\cdot)$ is the regularized standard deviation given by
\begin{equation}
    S(x, \epsilon) = \sqrt{\mathrm{Var}(x) + \epsilon}.
\end{equation}
This term enforces a minimum level of variability for each embedding dimension within the batch, thereby explicitly preventing representation collapse.
We use $1e-4$ as $\epsilon$ in our experiments.
The invariance regularization term is defined as follows:
\begin{equation}
    s(Z, Z') = \frac{1}{n} \sum_{i=1}^n {\| z_{i,:} - z_{i,:}'\|}_2^2.
\end{equation}
This term encourages the embeddings of two missing patterns to be close in the embedding space.
The covariance regularization term is defined as
\begin{equation}
    c(Z) = \frac{1}{l} \sum_{\substack{1 \leq i,j \leq l \\ i \neq j}} ( [C(Z)]_{i,j} )^2,
\end{equation}
where $C(Z)$ is the variance-covariance matrix of the embeddings:
\begin{equation}
    C(Z) = \frac{1}{n-1} \sum_{i=1}^n {(z_{i,:} - \bar{z}_{,:})^T(z_{i,:} - \bar{z}_{,:})},
\end{equation}
and $\bar{z}_{,:}$ is the mean embedding vector over the batch.
This term drives the off-diagonal elements of the variance-covariance matrix toward zero, reducing feature redundancy and ensuring that each dimension of the embedding space is effectively utilized.
The overall loss function is a weighted sum of the three components:
\begin{equation}
\begin{split}
    \mathcal{L}(Z, Z') &= 
    \lambda [v(Z) + v(Z')] \\
    &\quad + \mu [s(Z, Z')] 
    + \nu [c(Z) + c(Z')],
\end{split}
\end{equation}
where $\lambda$, $\mu$, and $\nu$ are weighting coefficients.

By forcing representations derived from different subsets of available stations
to be aligned in the embedding space, the feature extractor is encouraged to
internalize cross-station correlations rather than relying on any specific station.
As a result, the learned representation becomes invariant to station-wise feature missingness,
enabling robust inference even when a subset of stations is unavailable.

\subsection{Downstream Model Training with SMA}\label{subsec:downstream}
While CroSSL enables the feature extractor to learn representations invariant to station-wise
feature missingness, this alone is insufficient for robust end-to-end inference.
During downstream model training, the model is typically optimized using fully observed
multi-station inputs, whereas at inference time, CSI from some stations may be unavailable.
This mismatch between training and inference conditions can lead to performance degradation.

To mitigate this gap, we introduce Station-wise Masking Augmentation (SMA) during downstream
model training.
Given a labeled multi-station CSI input $X_i = [X_i^{(1)}, \ldots, X_i^{(N_d)}]$,
a masking set $\mathcal{M}_i^{\mathrm{mask}} \subseteq \{1, \ldots, N_d\}$ is sampled,
and the augmented input $\widetilde{X}_i$ is defined as
\begin{equation}
\widetilde{X}_i^{(d)} =
\begin{cases}
X_i^{(d)}, & \text{if } d \notin \mathcal{M}_i^{\mathrm{mask}}, \\
M_{\mathrm{inp}}, & \text{if } d \in \mathcal{M}_i^{\mathrm{mask}}.
\end{cases}
\end{equation}

Similar to the pre-training stage, the masking set $\mathcal{M}_i^{\mathrm{mask}}$
is determined by a masking probability $p_{\mathrm{mask}}$.
Each station is independently masked with probability $p_{\mathrm{mask}}$:
\begin{equation}
\mathbb{P}(d \in \mathcal{M}_i^{\mathrm{mask}}) = p_{\mathrm{mask}}.
\end{equation}

The augmented input $\widetilde{X}_i$ is then processed by the pre-trained encoders and
aggregator to obtain a global embedding:
\begin{equation}
Z_i = A\big( E^{(1)}(\widetilde{X}_i^{(1)}), \ldots, E^{(N_d)}(\widetilde{X}_i^{(N_d)}) \big).
\end{equation}
The downstream model $D(\cdot)$ produces a prediction $\hat{Y}_i = D(Z_i)$,
and is trained by minimizing a task-specific supervised loss $\ell(\cdot)$:
\begin{equation}
\mathcal{L}_{\mathrm{sup}} = \ell(\hat{Y}_i, Y_i).
\end{equation}

By applying station-wise masking at the input level during downstream training, the model explicitly learns to perform inference under incomplete multi-station observations.
Importantly, SMA complements CroSSL by transferring missingness robustness from the representation learning stage to the task-specific prediction stage.



\section{Evaluation using office-like environment dataset}\label{sec:evaluation_office}
To verify the effectiveness of the proposed method, we conducted experiments using two datasets: office-like environment dataset and factory-like environment dataset.
This section focuses on the evaluation using the office-like environment dataset.
CSI and RGB images were collected while a single subject walked in a standard office environment.
Informed consent was obtained from the participant prior to data collection.
Using the collected data, we constructed a one-dimensional position estimation dataset and performed machine learning experiments to evaluate our proposed framework's robustness to station-wise feature missingness and limited labeled data.

\begin{figure*}[]
  \centering
  \begin{subfigure}[b]{0.32\linewidth}
    \centering
    \includegraphics[width=\linewidth]{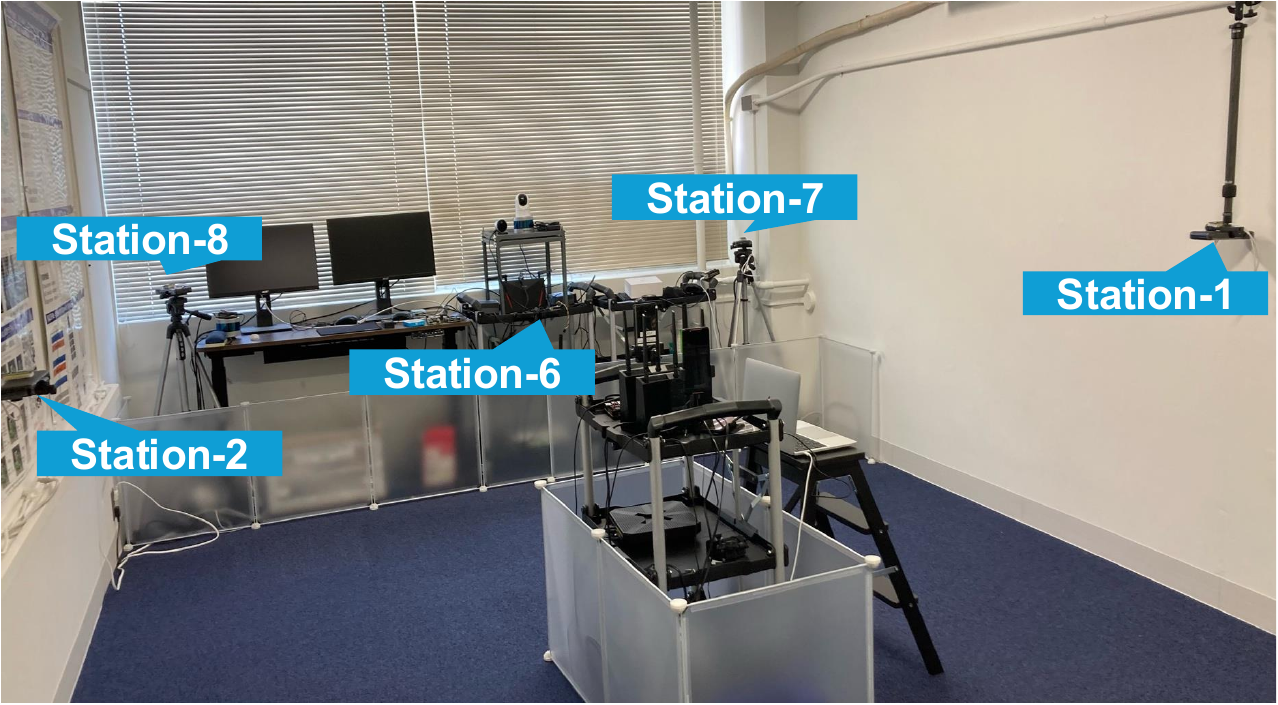}
    \caption{Snapshot 1.}
  \end{subfigure}
  \begin{subfigure}[b]{0.32\linewidth}
    \centering
    \includegraphics[width=\linewidth]{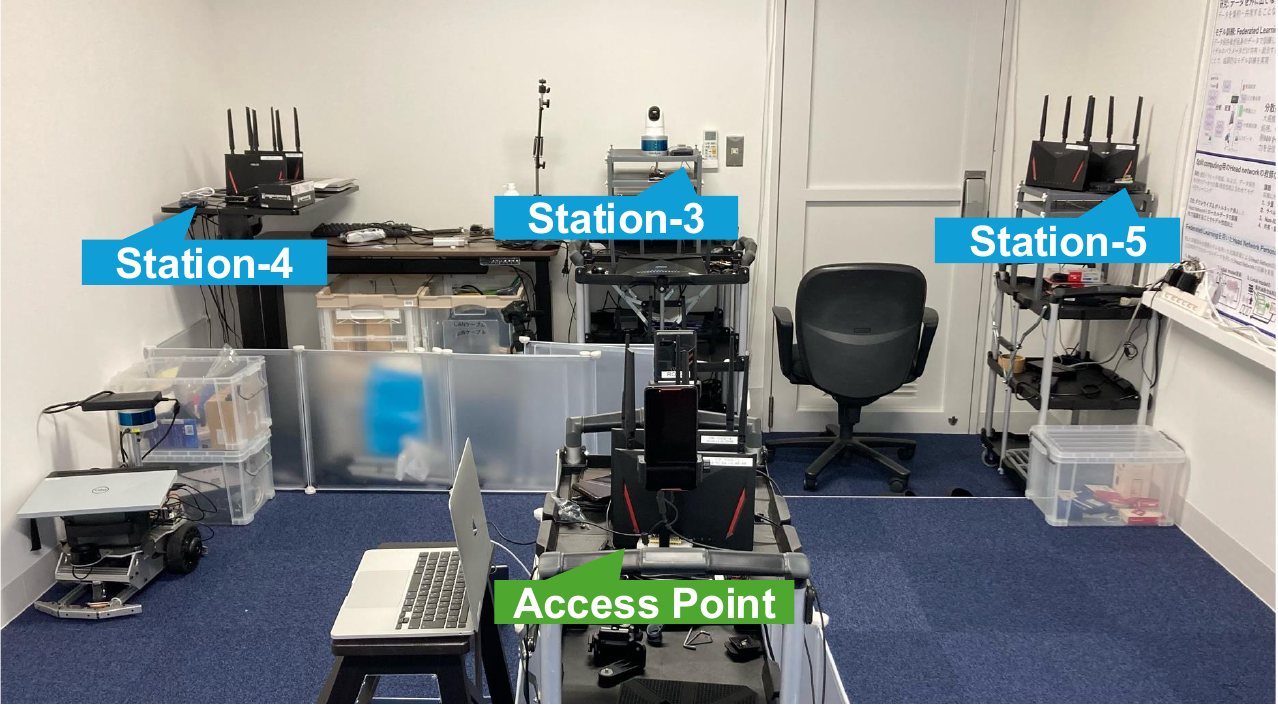}
    \caption{Snapshot 2.}
  \end{subfigure}
  \begin{subfigure}[b]{0.32\linewidth}
    \centering
    \includegraphics[width=\linewidth]{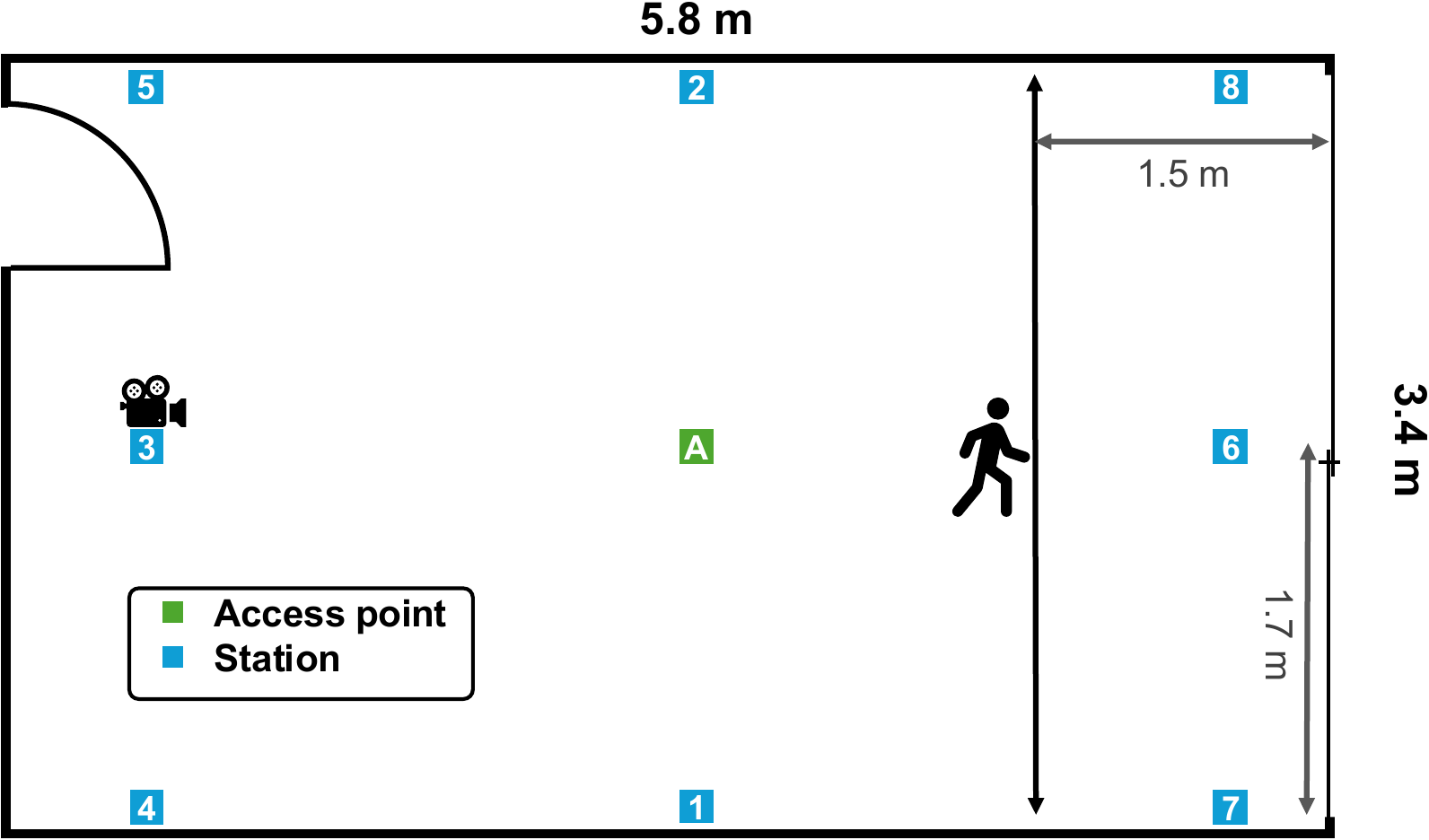}
    \caption{Layout}
  \end{subfigure}
  \caption{Experimental setup of office-like environment.}
  \label{fig:office_setup}
\end{figure*}

\begin{table*}[tb]
    \centering
    \caption{Experimental configuration}
    \begin{tabular}{lcc}
    \toprule
         & office-like environment  & factory-like environment \\ 
    \midrule
    AP & ESP32 &  NETGEAR Nighthawk X 10 \\
    Station & ESP32  & Raspberry Pi 4 model B \\
    Wireless LAN standard & 802.11n & 802.11ac \\
    Channel & 6 & 36 \\
    Bandwidth & 20\,MHz & 80\,MHz \\
    \midrule
    CSI sensor firmware & ESP32-CSI-Tool \cite{esp32:csitool} & Nexmon CSI \cite{nexmon:project, nexmon:paper} \\
    CSI measurement rate & 20\,Hz & 500\,Hz \\
    \midrule
    Camera & Macbook Pro Built-in FaceTime HD Camera & RealSense L515\\
    Camera measurement rate & 30\,Hz & 5\,Hz \\
    \bottomrule
    \end{tabular}
    \label{tab:dataset}
\end{table*}

\begin{table}[]
    \small
    \centering
    \setlength{\tabcolsep}{4pt}
    \caption{Optimized hyperparameters}
    \label{tab:optimized_hyperparams}
    \begin{tabular}{ccc}
        \toprule
        \textbf{Hyperparameter} & \textbf{office-like env} & \textbf{factory-like env} \\
        \midrule
        \multicolumn{3}{l}{\textbf{Downstream Model}} \\
        \hspace{5mm} learning rate & - & 2.0e-5 \\
        \midrule
        \multicolumn{3}{l}{\textbf{Pre-training: AutoFi}} \\
        \hspace{5mm} learning rate & 6.0e-4 & 2.4e-4  \\
        \hspace{5mm} noise scale & 4.5e-2 & 9.0e-5 \\
        \hspace{5mm} probability size & 16 & 32  \\
        \hspace{5mm} $\alpha_m$ & 2.3 & 1.5 \\
        \hspace{5mm} $\alpha_g$ & 2.1 & 92 \\
        \midrule
        \multicolumn{3}{l}{\textbf{Pre-training: DAE}} \\
        \hspace{5mm} learning rate & 1.6e-4 & 6.6e-3    \\
        \midrule
        \multicolumn{3}{l}{\textbf{Pre-training: CroSSL}} \\
        \hspace{5mm} learning rate & 1.1e-4 & 7.8e-4   \\
        \hspace{5mm} $\lambda$ & 5.4 & 69   \\ 
        \hspace{5mm} $\mu$ & 34 & 1.2e-2  \\ 
        \hspace{5mm} $\nu$ & 1.4e-2 & 7.4e-3  \\ 
        \bottomrule
    \end{tabular}
\end{table}

\subsection{Setup}\label{subsection:setup_office}
\noindent\textbf{Experimental setup for multi-station CSI data acquisition:}
The experimental setup and layout are illustrated in Fig.~\ref{fig:office_setup}.
A single AP was placed at the center of the room, and eight stations were positioned around the edge. 
One pedestrian walked along the trajectories indicated by the black arrows, introducing temporal variations in the wireless propagation environment. 
Table~\ref{tab:dataset} summarizes the hardware configuration and data acquisition settings.

Each station sequentially sent frames to the AP, and the AP measured CSI from the received frames. 
CSI acquisition was performed using the ESP32-CSI-Tool\cite{esp32:csitool}. 
Over a 20-minute experiment in the 2.4\,GHz band with IEEE 802.11n, single-antenna CSI of size $1 \times 1 \times 64$ was collected at an acquisition rate of approximately $R_{\mathrm{CSI}} \approx 20 \,\mathrm{Hz}$. 
The AP was connected to a laptop, and timestamps were added upon CSI acquisition. 

RGB images were captured at approximately $R_{\mathrm{Label}} \approx 30\,\mathrm{Hz}$ using the built-in FaceTime HD camera of a MacBook Pro.
The laptops recording CSI and RGB images were synchronized via Apple’s time server to ensure timestamp alignment.
From the captured images, supervision signals were generated depending on the downstream task.

\vspace{1mm}
\noindent\textbf{Downstream task:}
In this dataset, a one-dimensional position estimation task was considered.
Human positions were generated by applying the YOLO object detection algorithm \cite{yolo11} to the captured images. 
The bounding box of the detected person was extracted from each frame, and the normalized x-coordinate of its center was used as the ground-truth label. 
If object detection failed in a frame, the missing label was interpolated by averaging the center coordinates of the two adjacent frames. 
As a result, the normalized labels ranged from approximately 0.166 to 0.854, corresponding to an actual horizontal position range of 0 to 3.4 meters in the experimental setup.

The collected CSI was preprocessed following the standard procedure described in Subsection~\ref{subsec:assumption}, resulting in real-valued vectors of size $1 \times 1 \times 52$.
Because both the transmitter and receiver are equipped with a single antenna, the antenna dimensions do not provide additional spatial diversity.
Thus, the CSI was represented as a $52$-dimensional real-valued vector.

\vspace{1mm}
\noindent\textbf{Dataset construction:}
From the preprocessed CSI and supervision signals, labeled and unlabeled samples were generated following the procedure described in previous section.
The 20\,minute dataset was split along the time axis into training, validation, and test sets with a ratio of 7:1.5:1.5.
Unlabeled samples were generated at $R_{\mathrm{SSL}} = 160\,{\mathrm{Hz}}$ with a temporal window $w = 2.0\,\mathrm{s}$, resulting in a total of 134,400 unlabeled samples. Among them, 134,234 samples contained CSI from all eight stations, and 166 samples contained CSI from seven stations. 
Labeled samples were generated using $R_{\mathrm{Label}} = 30\,\mathrm{Hz}$ with the same temporal window. 
As a result, 25,200, 5,400, and 5,400 labeled samples were obtained for training, validation, and testing, respectively. 
Among the training samples, 25,168 contained CSI from all eight stations, and 32 contained CSI from seven stations. 
For the validation set, 5,369 samples contained CSI from all eight stations, whereas 31 samples contained CSI from seven stations.
All 5,400 test samples contained CSI from all eight stations.

Although the collected data rarely contained missing stations, station-wise feature missingness was explicitly simulated during training and evaluation.
Specifically, during pre-training and downstream model training, station-wise masking was applied according to the proposed framework to emulate realistic station unavailability scenarios.
At inference time, performance under different numbers of available stations was evaluated by masking the corresponding station inputs.

\vspace{1mm}
\noindent\textbf{Model configuration and training details:}
The constructed datasets were then used to train and evaluate machine learning models. 
For model development, the proposed framework allows flexible selection of the feature extractor and downstream model depending on the task.
For the office-like environment dataset, each station’s CSI was processed by an identity encoder without learnable parameters, as the CSI inputs were already temporally aggregated.
The resulting station-wise features were then combined by an aggregator consisting of three repeated blocks of a dense layer, ReLU activation, batch normalization, and dropout, with a dropout rate of 0.3.
The size of the feature extractor was approximately 2.7\,MB when stored in single-precision floating-point format.

The downstream model was implemented using a RandomForestRegressor.
As this downstream model is not trained via gradient-based backpropagation, it cannot be jointly fine-tuned with the neural network-based feature extractor. 
Therefore, the feature extractor's weights were frozen, and the downstream model was trained using the features generated by the pre-trained feature extractor.
SSL was performed using TensorFlow 2.19 with the Adam optimizer, while downstream training used scikit-learn 1.7.0.
The batch size for pre-training was set to 4096.
Early stopping was applied when the training loss did not improve for 10 consecutive epochs, with a maximum of 1000 epochs.
Hyperparameter optimization was conducted using Optuna 4.4 \cite{optuna} with its default optimization algorithm, performing 30 trials per hyperparameter set and selecting the best-performing configuration.
The final hyperparameter values used for training are summarized in Table~\ref{tab:optimized_hyperparams}.
To ensure the reliability of the results, all experiments were repeated three times with different random seeds.

\subsection{Baseline}\label{subsec:baseline_office}
To evaluate the effectiveness of the proposed method, we introduce several baseline methods for performance comparison. 
These baselines are designed to assess the impact of pre-training and data augmentation techniques.

\vspace{1mm}
\noindent\textbf{Constant:}
We introduce Constant as the simplest non-learning baseline. 
This method continuously predicts a fixed value. 
For the one-dimensional position estimation task, it consistently predicts 0.5, which represents the center coordinate of the trajectory. This approach serves as a baseline to indicate the inherent difficulty of the task.

\vspace{1mm}
\noindent\textbf{Fully Supervised Baselines:}
As fully supervised baselines, we introduce two methods trained solely on labeled data. 
The first is NaiveSupervised.
For the office-like environment dataset, this baseline fed the CSI, concatenated along the station dimension, directory into the downstream model, bypassing any feature extractor.
The second is an OutputEnsemble, in which $N_d$ separate downstream models are trained, and their outputs are averaged during inference. 
These methods serve as reference points for evaluating performance without pre-training.

\vspace{1mm}
\noindent
\textbf{Pre-training Baselines:}
We evaluate three pre-training baselines—AutoFi~\cite{autofi}, Denoising AutoEncoder (DAE) and CroSSL~\cite{crossl}—in which a feature extractor is first pre-trained using unlabeled data and subsequently applied to the downstream task.

For AutoFi, we tune the learning rate, the variance of the additive Gaussian noise used for data augmentation (noise scale), the dimensionality of the probability space used in the loss computation, and the weighting coefficients $\alpha_m$ and $\alpha_g$ for the mutual information loss and geometric consistency loss, respectively.
The feature extractor architecture of AutoFi is identical to the aggregator network used in the proposed method, ensuring a fair comparison.

DAE learns representations by reconstructing clean inputs from corrupted versions using an encoder–decoder architecture.
During downstream task training, only the encoder is retained and used as the feature extractor.
Denoising AutoEncoders have been shown to be effective for handling missing values in related domains, such as power load forecasting~\cite{DenoisingAE}.
In our implementation, Station-wise Masking is applied to the input CSI, and the model is trained to reconstruct the original CSI.
The loss function is defined as the mean squared error between the original and reconstructed values over the masked portions.
The encoder architecture of DAE is identical to the aggregator used in the proposed method.
The decoder is implemented as a symmetric but shallower network composed of two dense layers with batch normalization and dropout.
Unless otherwise specified, we set the masking ratio to $p_{\mathrm{mask}} = 0.5$ for both DAE and CroSSL.

\vspace{1mm}
\noindent
\textbf{Augmentation Baselines:}
To examine the effect of data augmentation, we apply three augmentation methods—Random Erasing (RE), Station-wise Masking Augmentation (SMA) and Inpainting—to the NaiveSupervised baseline.

The RE method is inspired by object-aware erasing~\cite{RandomErasing} and adapted to multi-station CSI data.
Specifically, for each station, a continuous region along the subcarrier dimension is independently masked.
The erased region length is randomly sampled from the range $[s_l, s_h] = [0.4, 0.6]$ for each sample.

For SMA, station-wise masking is applied with a masking probability of $p_{\mathrm{mask}} = 0.5$.
In the office-like environment dataset, an offline augmentation strategy is adopted: each training sample is augmented once, and the augmented samples are added to the training set, effectively doubling its size.

For the Inpainting baseline, missing portions of the CSI are reconstructed using a Context Encoder-based \cite{contextencoder} inpainting model, following the idea proposed in \cite{recover_multiCSI}.
In our implementation, we use the same encoder-decoder architecture and training procedure as the DAE baseline, but explicitly apply it to recover missing CSI before downstream inference.
Unlike \cite{recover_multiCSI}, which leverages temporal continuity within each station, our setting restricts reconstruction to cross-station information at the same time instance, making the recovery problem substantially more challenging.

\begin{figure}[]
  \centering
  \begin{subfigure}[b]{\linewidth}
    \centering
    \includegraphics[width=\linewidth]{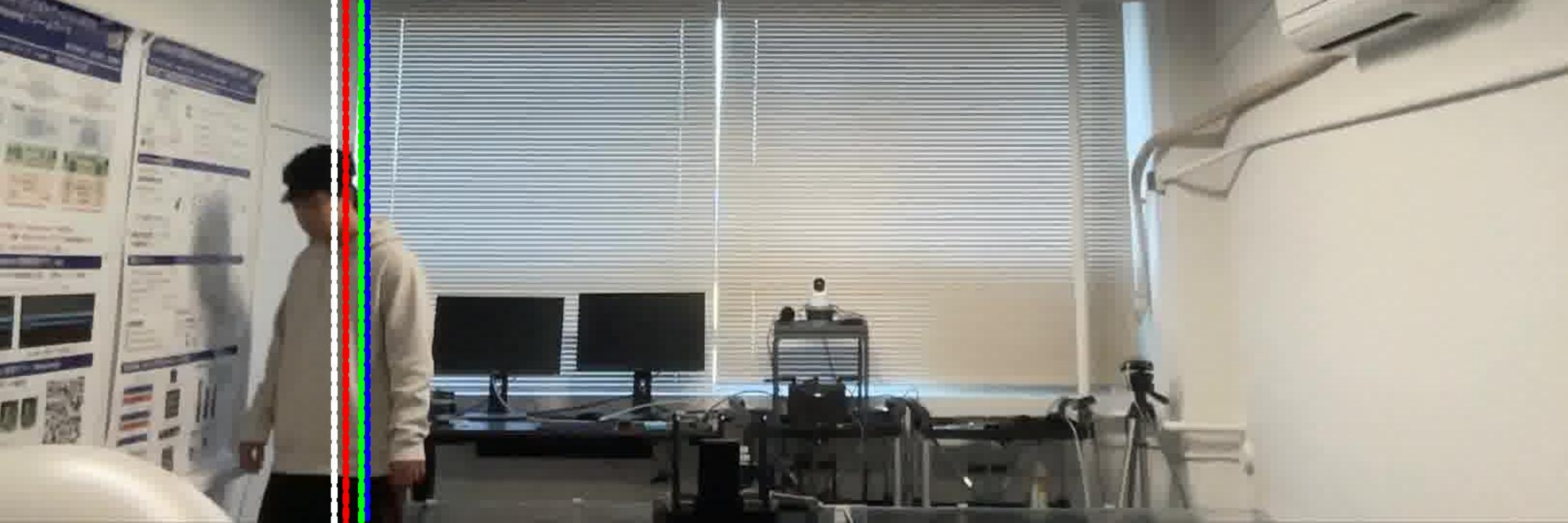}
    \caption{Sample1: $\mathrm{RMSE}_{1/4/8\,\mathrm{stations}} = 0.0212 / 0.0173 /  0.0074 $}
  \end{subfigure}
  \begin{subfigure}[b]{\linewidth}
    \centering
    \includegraphics[width=\linewidth]{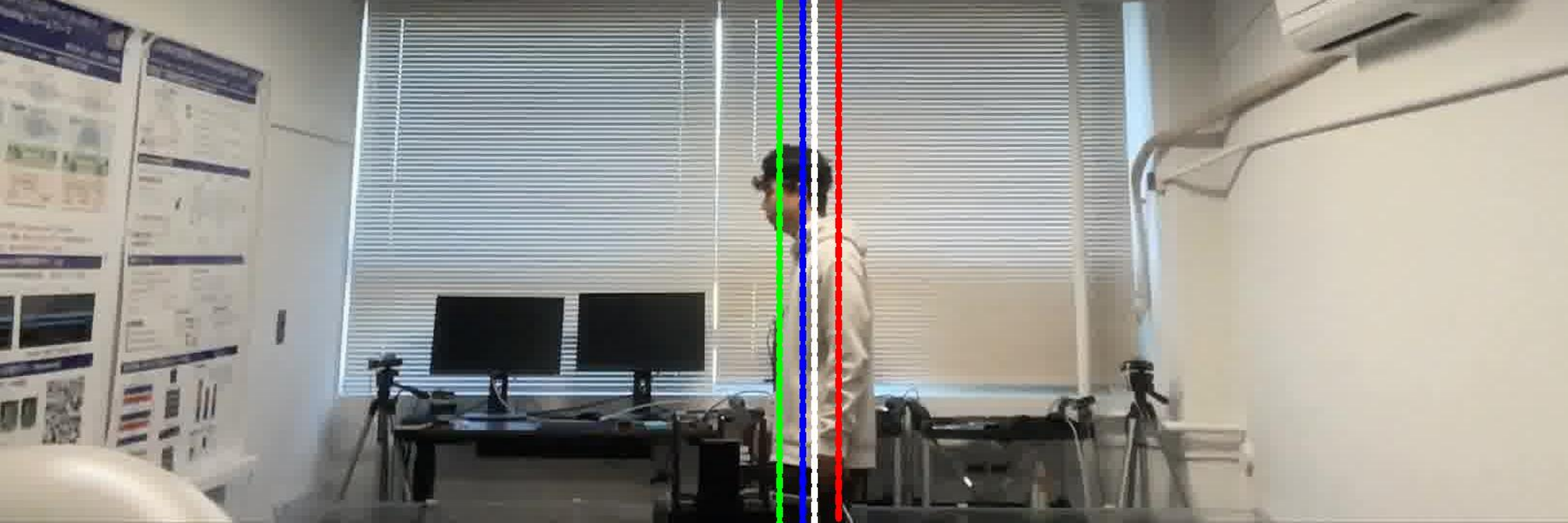}
    \caption{Sample2: $\mathrm{RMSE}_{1/4/8\,\mathrm{stations}} = 0.0074 / 0.0225 / 0.0154 $}
  \end{subfigure}
  \begin{subfigure}[b]{\linewidth}
    \centering
    \includegraphics[width=\linewidth]{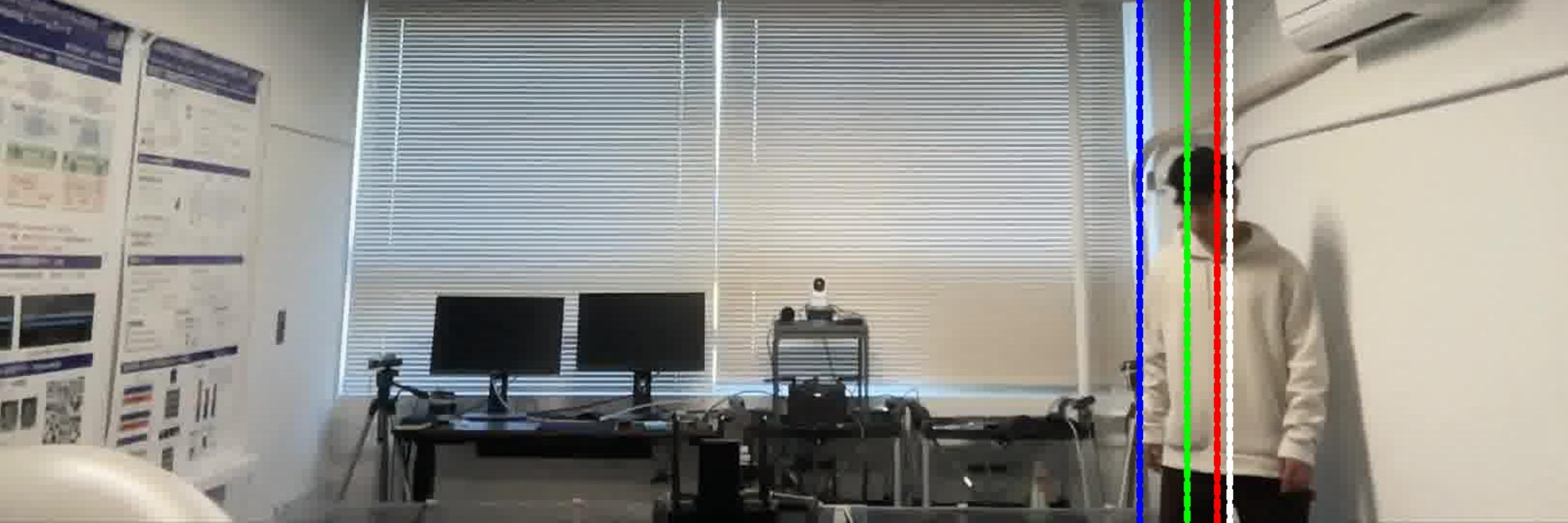}
    \caption{Sample3: $\mathrm{RMSE}_{1/4/8\,\mathrm{stations}} = 0.0570 / 0.0271 / 0.0077$}
  \end{subfigure}
  \caption{ Examples of one-dimensional position estimation. The white line denotes the ground-truth label, while the blue, green, and red lines indicate the predictions obtained with 1, 4, and 8 stations, respectively.
  $\mathrm{RMSE}_{x\,\mathrm{stations}}$ represents the RMSE of using $x$\,stations.}
  \label{fig:at_a_glance_office}
\end{figure}

\subsection{At a Glance}
To qualitatively assess the robustness of proposed method to station-wise feature missingness, we visualize examples of the predicted positions under different missingness conditions.
Fig.~\ref{fig:at_a_glance_office} illustrates the predicted positions obtained by the proposed method when trained with all labeled data.
The white line denotes the ground-truth, while the blue, green and red lines represent the predictions obtained with 1, 4 and 8 stations at inference.
Here, "with $\mathrm{n}$ stations at inference" means that CSI from only $\mathrm{n}$ stations are provided to the model, while the remaining stations are treated as missing.
Specifically, the CSI of unavailable stations are masked with zero.
In this visualization, fixed station combinations are used for clarity.
Station-6 is used for the 1-station case, Stations-1, 2, 3 and 6 are used for the 4-stations case, and all stations are used for the 8-stations case.
Note that this figure shows a representative example with fixed station combinations, whereas the quantitative evaluations average results over all possible station combinations.
Across the three samples, the proposed method maintains its performance even when the number of available stations is reduced by half.
Notably, the increase in prediction error is only marginal even with a single station.
These observations suggest that the proposed method is robust to station-wise feature missingness.


\begin{figure}
    \centering
    \includegraphics[width=\linewidth]{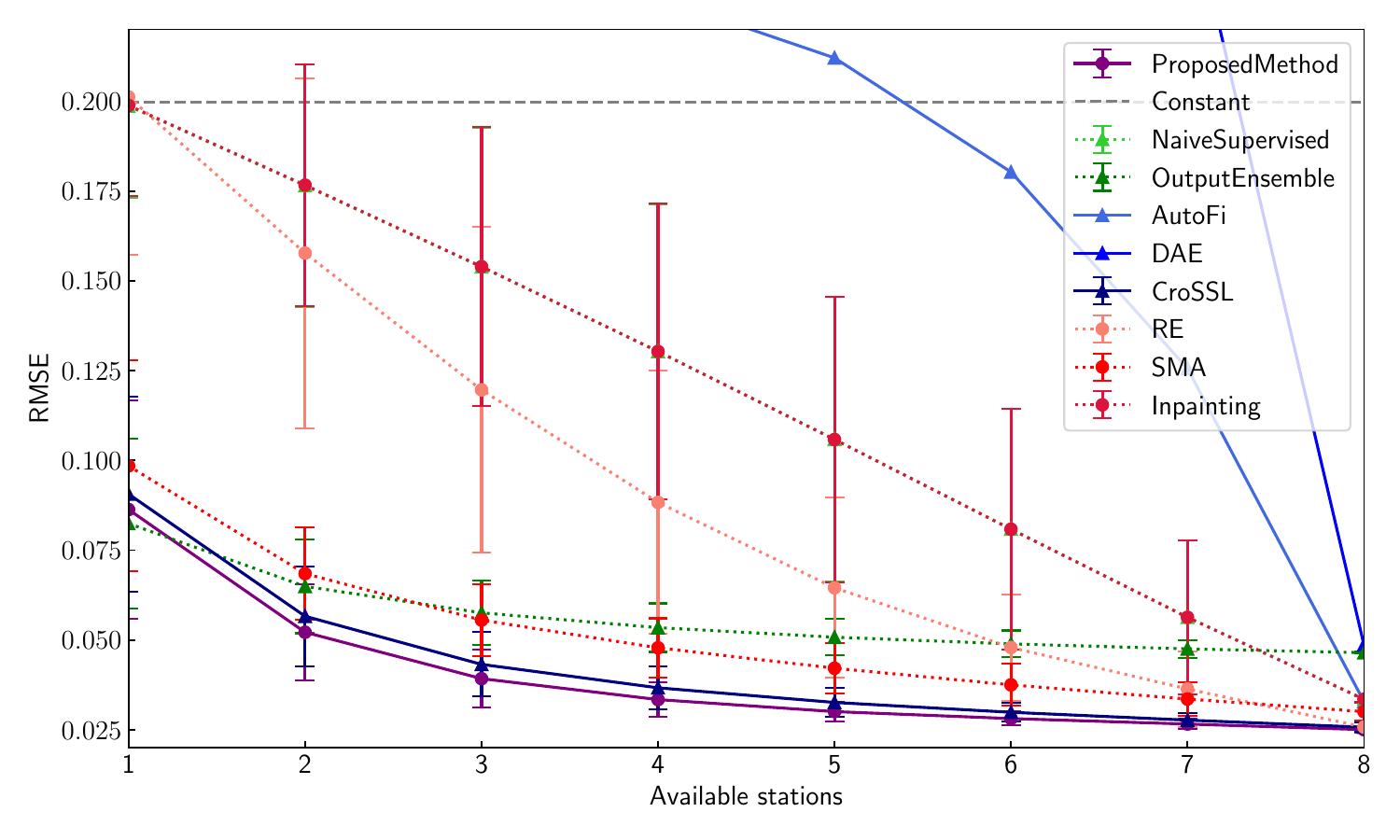}
    \caption{
    Performance comparison under different numbers of available stations on the office-like environment dataset.
    The proposed method is shown as the purple solid line.
    For methods whose worst-case RMSE exceeds 0.21, the standard deviation bars are omitted for clarity.
    While most baselines degrade as stations become unavailable,
    the proposed method maintains low RMSE, demonstrating robustness to station-wise feature missingness.
    }
    \label{fig:evalu_missing_device_office}
\end{figure}
\subsection{Robustness to Station-wise Feature Missingness}
Following the qualitative examples presented in the previous subsections, we now provide a quantitative evaluation of robustness to station-wise feature missingness.
Fig.~\ref{fig:evalu_missing_device_office} summarizes the performance of each method when varying the number of available stations at inference time, while all models are trained with 100\% of the labeled samples.

The proposed method consistently maintains the lowest error across a wide range of missingness levels.
Even when more than half of the stations are unavailable, its performance degrades only marginally, demonstrating strong robustness to station-wise feature missingness.
This indicates that the proposed method does not rely heavily on any particular station and retains its discriminative power even under substantial feature loss.

Moreover, the proposed method achieves low error not only under severe station-wise feature missingness but also when all stations are available.
Unlike OutputEnsemble, which maintains robustness by isolating each station but fail to exploit cross-station correlations, the proposed method learns a unified representation that integrates complementary information across stations.
As a result, it avoids the performance saturation observed in ensemble methods while retaining robustness to missing stations.

These results collectively demonstrate that the proposed method simultaneously leverages multi-station information when available and remains robust when stations are missing, making it well-suited for real-world deployments where station-wise feature missingness is unavoidable.

\begin{figure}
    \centering
    \includegraphics[width=\linewidth]{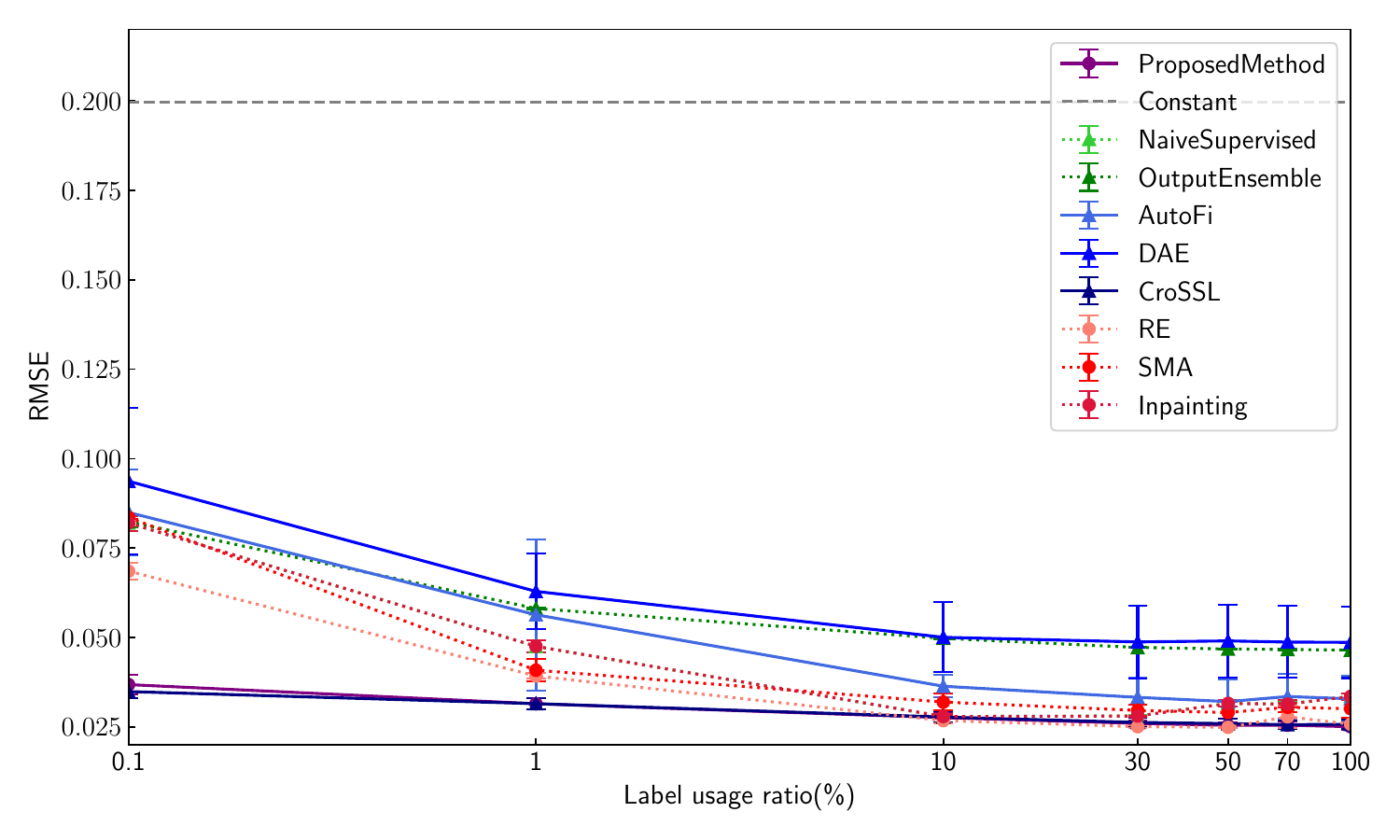}
    \caption{Performance comparison under different amounts of labeled data used for downstream model training on the office-like environment dataset.
     While the performance of most baselines degrades as the amount of labeled data decreases, the proposed method maintains stable performance and remains competitive across all label ratios, demonstrating robustness to limited labeled data.
    }
    \label{fig:eval_limited_label_office}
\end{figure}
\subsection{Robustness to Limited Labeled Data}
After evaluating robustness to station-wise feature missingness, we next investigate how each method behaves when labeled data is scarce.
Fig.~\ref{fig:eval_limited_label_office} summarizes the performance of each method when varying the amount of labeled data used for downstream model training in the office-like environment.

The proposed method and CroSSL maintain stable performance even when the amount of labeled data is severely limited.
This robustness stems from self-supervised pre-training, which enables the model to learn informative representations from unlabeled CSI before downstream training.

In comparison, fully supervised and augmentation-based baselines achieve competitive performance only when sufficient labeled data is available, but their error increases rapidly as the label usage ratio decreases.
This contrast highlights that effective pre-training is essential, especially in low-label regimes.

\begin{table}[]
\caption{RMSE comparison under different numbers of available stations and label ratios on the office-like environment dataset. 
For each condition, the reported RMSE is averaged over all combinations of station-missingness patterns.
Best results for each condition are shown in bold.
}
\centering

\begin{tabular}{@{} l *{9}{c} @{}}
    \toprule
    Label usage ratio(\%)
    & \multicolumn{2}{c}{0.1}
    & \multicolumn{2}{c}{10} \\
    \cmidrule(lr){2-3}\cmidrule(lr){4-5}
    Available stations& 1 & 4 & 1 & 4  \\
    \midrule
    ProposedMethod & $0.0924$ & $0.0424$ & $0.0867$ & $\mathbf{0.0353}$ \\
    \cdashline{1-5}
    Constant & $0.1977$ & $0.1977$ & $0.1977$ & $0.1977$   \\ 
    NaiveSupervised & $0.1888$ & $0.1488$ & $0.1910$ & $0.1177$ \\
    OutputEnsemble  & $0.1233$ & $0.0894$ & $\mathbf{0.0845}$ & $0.0563$ \\
    \cdashline{1-5}
    AutoFi   & $0.2415$ & $0.2190$ & $0.2522$ & $0.2254$
    \\
    DAE      & $0.2201$ & $0.2271$ & $0.2381$ & $0.2460$
    \\
    CroSSL   & $\mathbf{0.0909}$ & $\mathbf{0.0403}$ & $0.0912$ & $0.0371$
    \\
    \cdashline{1-5}
    RE  & $0.2139$ & $0.1492$ & $0.2224$ & $0.0987$
    \\
    SMA & $0.2056$ & $0.1427$ & $0.1140$ & $0.0532$ 
    \\
    Inpainting & $0.1888$ & $0.1488$ & $0.1910$ & $0.1177$
    \\
    \bottomrule
\end{tabular}

\label{tab:office_combined_performance}
\end{table}

\subsection{Combined Robustness to Station-wise Feature Missingness and Limited Labeled Data}
We further evaluate the proposed method under the combined challenges of station-wise feature missingness and limited labeled data.
Table~\ref{tab:office_combined_performance} summarizes the RMSE under these joint conditions, allowing us to compare how each method behaves when both challenges arise simultaneously.

Across all conditions, the proposed method and CroSSL maintain consistently strong performance, even when both the number of available stations and the amount of labeled data are severely restricted. 
This indicates that pre-training using CroSSL enables the extraction of representations that are both robust to missing stations and effective for downstream tasks.
This dual robustness, against both station-wise feature missingness and label scarcity, highlights the practical value of the proposed method for deployment in environments where sensing conditions cannot be tightly controlled.

We note that the Inpainting baseline exhibits performance nearly identical to
NaiveSupervised under these combined conditions.
This suggests that explicit CSI reconstruction provides little additional benefit
in our setting, where missing CSI must be inferred solely from other stations at
the same time instance, without access to temporal continuity.
These results further support our design choice of learning station-missingness-
invariant representations, rather than relying on explicit CSI recovery.



\begin{figure}[]
  \centering

  \begin{subfigure}[b]{0.23\textwidth}
    \centering
    \includegraphics[width=\textwidth]{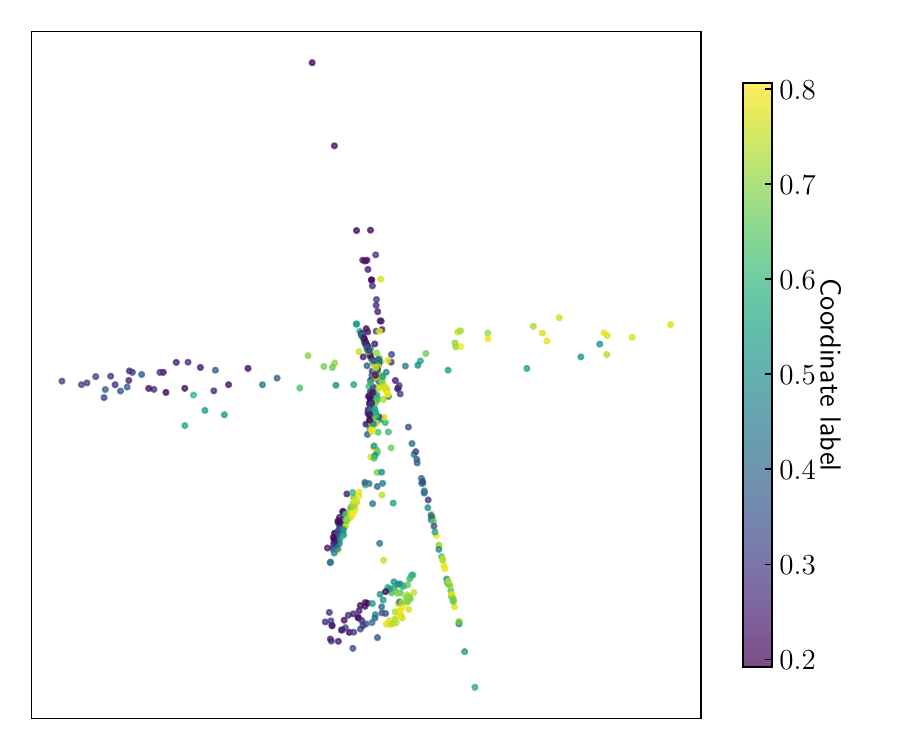}
    \caption{Input (1 station)}
    \label{subfig:visualize_input_1station}
  \end{subfigure}
  \hfill
  \begin{subfigure}[b]{0.23\textwidth}
    \centering
    \includegraphics[width=\textwidth]{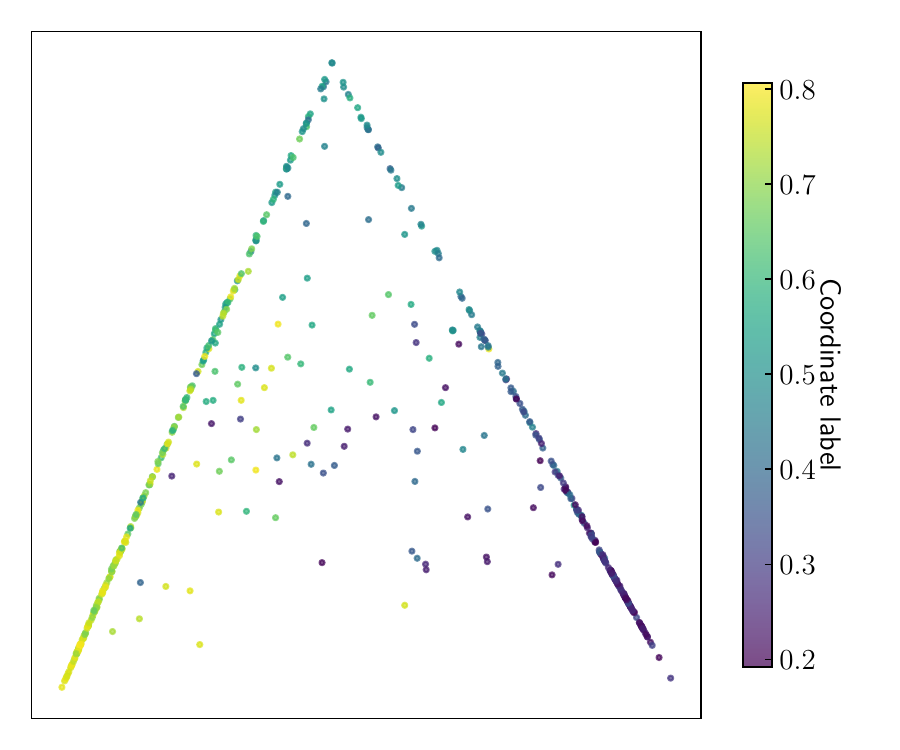}
    \caption{Embedding (1 station)}
    \label{subfig:visualize_embedding_1station}
  \end{subfigure}

  \vspace{2mm}
  \begin{subfigure}[b]{0.23\textwidth}
    \centering
    \includegraphics[width=\textwidth]{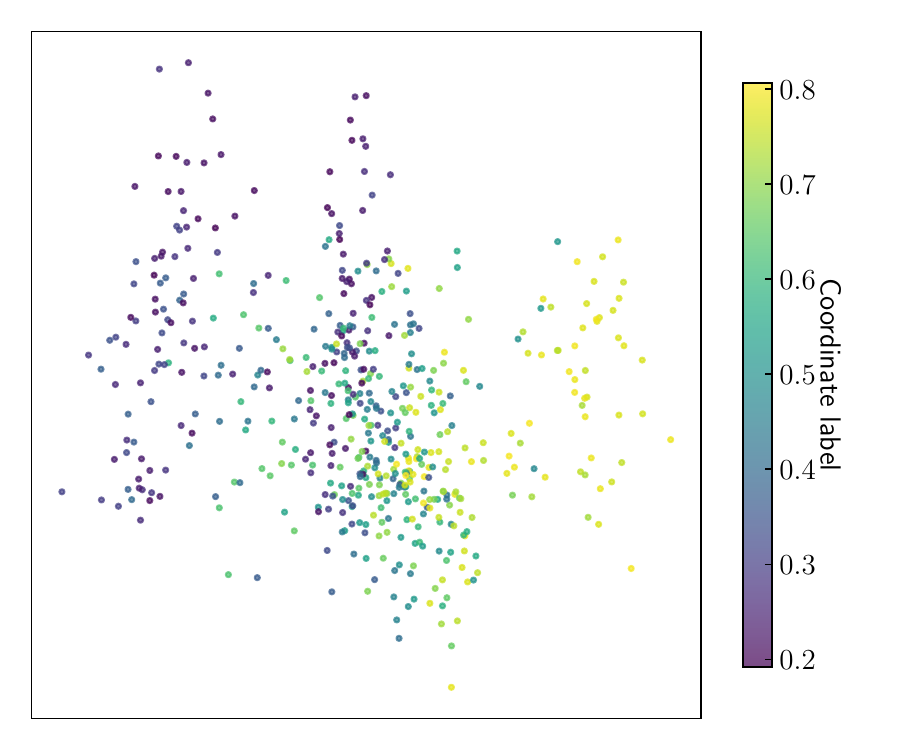}
    \caption{Input (4 stations)}
    \label{subfig:visualize_input_4station}
  \end{subfigure}
  \hfill
  \begin{subfigure}[b]{0.23\textwidth}
    \centering
    \includegraphics[width=\textwidth]{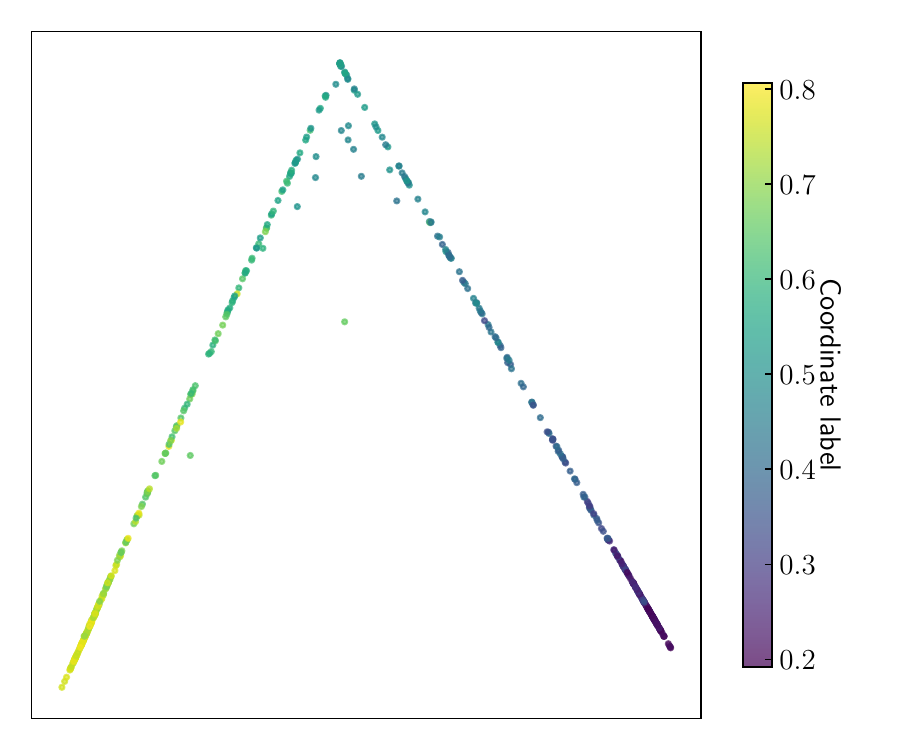}
    \caption{Embedding (4 stations)}
    \label{subfig:visualize_embedding_4station}
  \end{subfigure}

  \vspace{2mm}
  \begin{subfigure}[b]{0.23\textwidth}
    \centering
    \includegraphics[width=\textwidth]{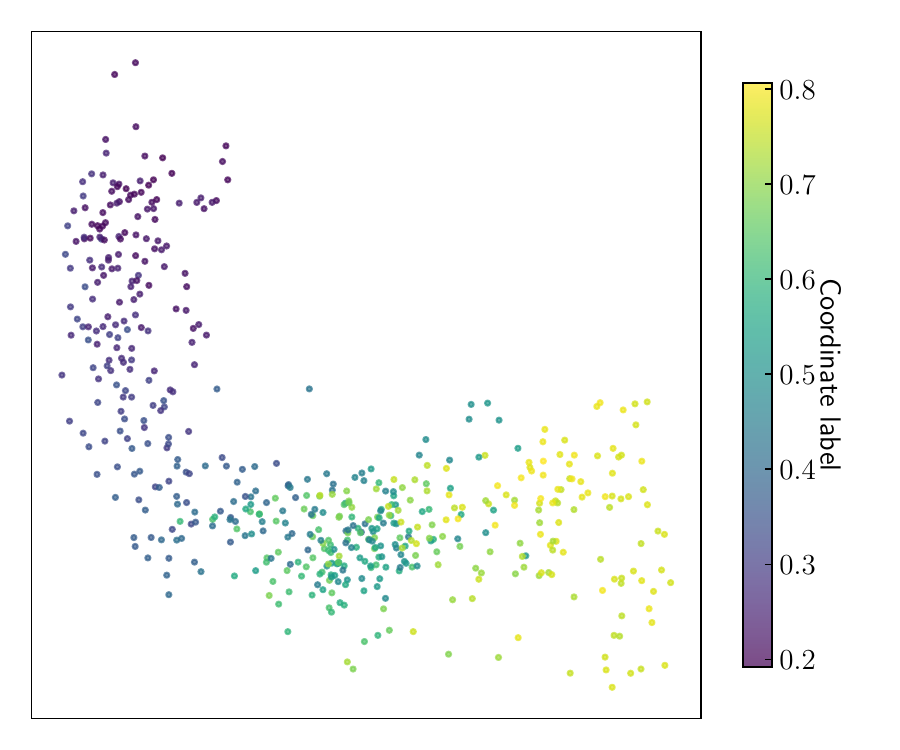}
    \caption{Input (8 stations)}
    \label{subfig:visualize_input_8station}
  \end{subfigure}
  \hfill
  \begin{subfigure}[b]{0.23\textwidth}
    \centering
    \includegraphics[width=\textwidth]{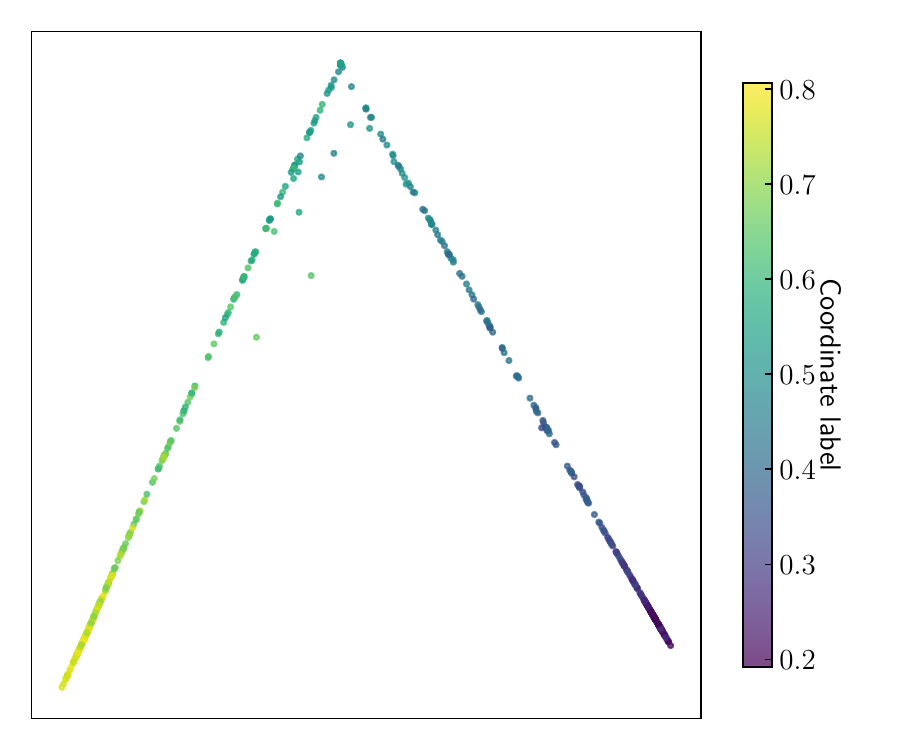}
    \caption{Embedding (8 stations)}
    \label{subfig:visualize_embedding_8station}
  \end{subfigure}

  \caption{
    PCA-transformed scatter plots of raw inputs and embeddings under different station-wise feature missingness conditions.
    Points colored by the ground-truth position.
    While the raw inputs become increasingly irregular as missingness increases, the embeddings maintain a smooth and consistent manifold, indicating robustness to station-wise feature missingness.
  }
  \label{fig:pca_visualizetion}
\end{figure}
\subsection{Visualization of Input and Embedding Spaces}
To understand why the proposed method achieves robustness to station-wise feature missingness, Fig.~\ref{fig:pca_visualizetion} visualizes the raw CSI inputs and the global embeddings learned by the proposed method under different numbers of available stations.
To obtain these visualizations, a PCA model is first fitted using the training set and then applied to the test set, where station-wise feature missingness is artificially induced.

When all stations are present, the raw CSI inputs form a smooth structure that aligns well with the ground-truth positions.
However, as station-wise feature missingness increases, the input distributions become irregular and begin to cluster according to missing patterns than the underlying labels, indicating that the raw CSI is highly sensitive to missing stations.

In contrast, the learned embeddings remain far more stable.
Although the embeddings become slightly more dispersed when fewer stations are available, the overall manifold structure—ordered consistently by the ground-truth position—remains preserved across all missingness conditions.
This demonstrates that the proposed method effectively projects inputs with different missing patterns into a shared manifold that retains task-relevant structure.

Overall, these results reinforce our claim of robustness to station-wise feature missingness:
while the raw CSI undergoes substantial distributional distortions due to feature missingness, the learned embeddings remain label-consistent, enabling reliable inference even under severe feature missingness.

\begin{table*}[]
\centering
\caption{RMSE comparison of different combinations of pre-training and data augmentation strategies on the office-like environment dataset under varying numbers of available stations and label ratios. 
For each condition, the reported RMSE is averaged over all combinations of station-missingness patterns.
Best results for each condition are in bold. 
The results show that the proposed method consistently achieves the lowest error.}

\begin{tabular}{@{} l *{9}{c} @{}}
    \toprule
    Label usage ratio(\%)
    & \multicolumn{3}{c}{0.1}
    & \multicolumn{3}{c}{10}
    & \multicolumn{3}{c}{100} \\
    \cmidrule(lr){2-4}\cmidrule(lr){5-7}\cmidrule(l){8-10}
    Available stations& 1 & 4 & 8 & 1 & 4 & 8 & 1 & 4 & 8 \\
    \midrule
    ProposedMethod & $\mathbf{0.0924}$ & $\mathbf{0.0424}$ & $\mathbf{0.0368}$ & $\mathbf{0.0867}$ & $\mathbf{0.0353}$ & $\mathbf{0.0276}$ & $\mathbf{0.0863}$ & $\mathbf{0.0334}$ & $\mathbf{0.0251}$  
    \\
    AutoFi × SMA & $0.2126$ & $0.1964$ & $0.0977$ & $0.1594$ & $0.1080$ & $0.0323$ & $0.1158$ & $0.0818$ & $0.0289$
    \\ 
    DAE × SMA & $0.2394$ & $0.2182$ & $0.1142$ & $0.1709$ & $0.1438$ & $0.0512$ & $0.1087$ & $0.0825$ & $0.0545$
    \\
    CroSSL × RE  &  $0.1083$ & $0.0670$ & $0.0619$ & $0.0940$ & $0.0391$ & $0.0285$ & $0.0926$ & $0.0388$ & $0.0267$     
    \\
    
    \bottomrule
\end{tabular}

\label{tab:office_comparison_combination}
\end{table*}
\subsection{Ablation Study of Pre-training and Augmentation Strategies}

The proposed method adopts CroSSL for pre-training and SMA for augmentation.
In this section, we evaluate the impact of replacing these components with alternative methods.
Table~\ref{tab:office_comparison_combination} summarizes the performance of different combinations of pre-training and augmentation strategies across various label ratios and numbers of available stations.

The proposed method achieved the best performance across all scenarios.
This confirms that combining missingness-invariant pre-training with station-aware augmentation improves robustness to station-wise feature missingness, and provides additional resilience under limited labeled data.

In addition, applying SMA to existing SSL methods such as AutoFi and DAE yields performance improvements, especially when fewer stations are available.
These results suggest that SMA acts as a simple yet powerful augmentation strategy for enhancing robustness to structured feature missingness.
Given its ease of integration, SMA can be incorporated into a wide range of multi-station CSI sensing frameworks to improve resilience to station-wise feature missingness without modifying the underlying model architecture.

\begin{figure}
    \centering
    \includegraphics[width=\linewidth]{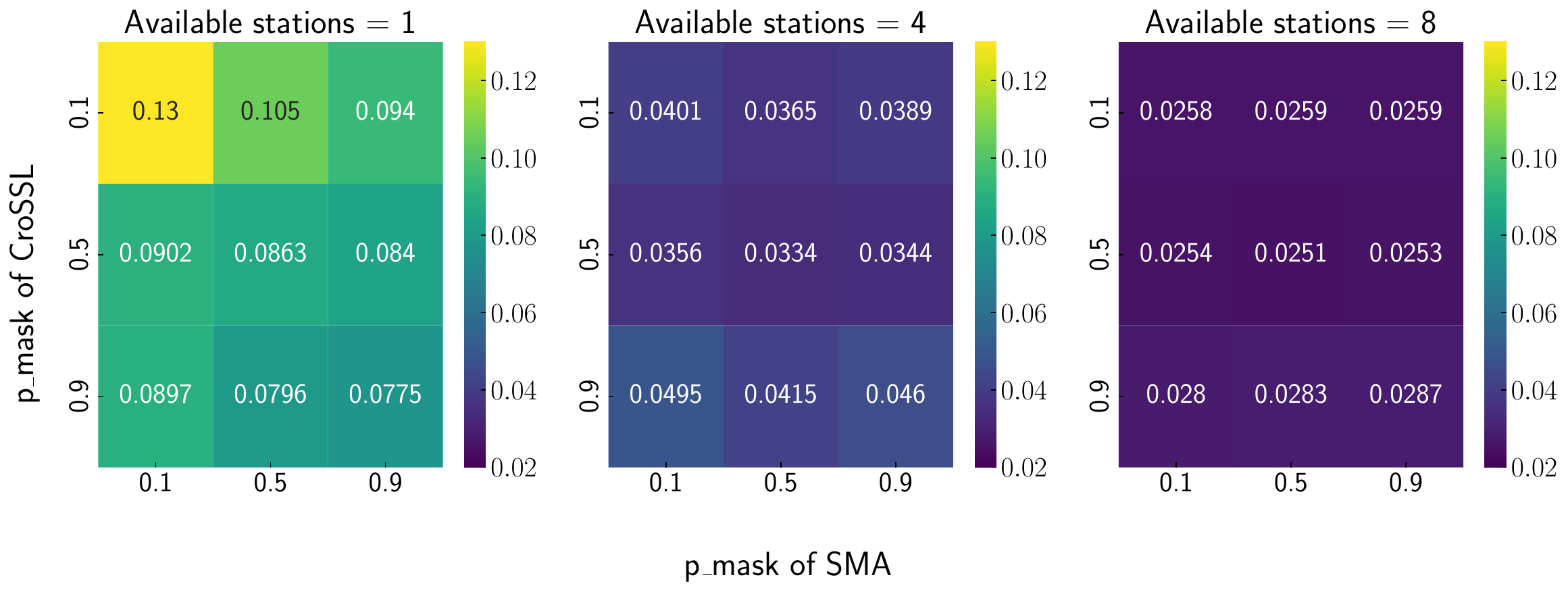}
    \caption{Each heatmap shows the RMSE for combinations of CroSSL and SMA $p_\mathrm{mask}$ values under different numbers of available stations on the office-like environment dataset. 
    Across all conditions, the combination $p_\mathrm{mask} = 0.5$ provides the most stable performance, suggesting that this masking rate offers an appropriate training difficulty for the office-like environment dataset.}
    \label{fig:masking_rate_office}
\end{figure}
\subsection{Sensitivity Analysis of Masking Rate}


In the previous experiments, the masking ratio $p_{\mathrm{mask}}$ used during both pre-training and downstream model training was fixed at 0.5.
This section investigates the effect of varying $p_{\mathrm{mask}}$ on model performance, using all labeled data.
Fig.~\ref{fig:masking_rate_office} summarizes the RMSE for combinations of CroSSL  $p_{\mathrm{mask}}$ and SMA  $p_{\mathrm{mask}}$ across different numbers of available stations at inference.

Overall, robust performances are consistently observed with $p_{\mathrm{mask}} = 0.5$.
Additionally, when only one station is available at inference time, performance also drops with $p_{\mathrm{mask}} = 0.1$ during both pre-training and downstream model training.
This can be attributed to the mismatch between the low missingness rate assumed during training and the high level of missingness encountered during inference.
These findings highlight that $p_{\mathrm{mask}}=0.5$ is a well-balanced choice for this dataset.

\section{Evaluation using factory-like environment dataset}

To further assess the robustness of proposed method, this section details an evaluation using factory-like environment dataset. 
The dataset was constructed by collecting CSI and RGB images as five subjects walked within a large, semi-enclosed indoor space. 
Informed consent was obtained from all participants prior to data collection.
From this dataset, we formulated an image generation task.

\begin{figure*}[]
  \centering
  \begin{subfigure}[b]{0.32\linewidth}
    \centering
    \includegraphics[width=\linewidth]{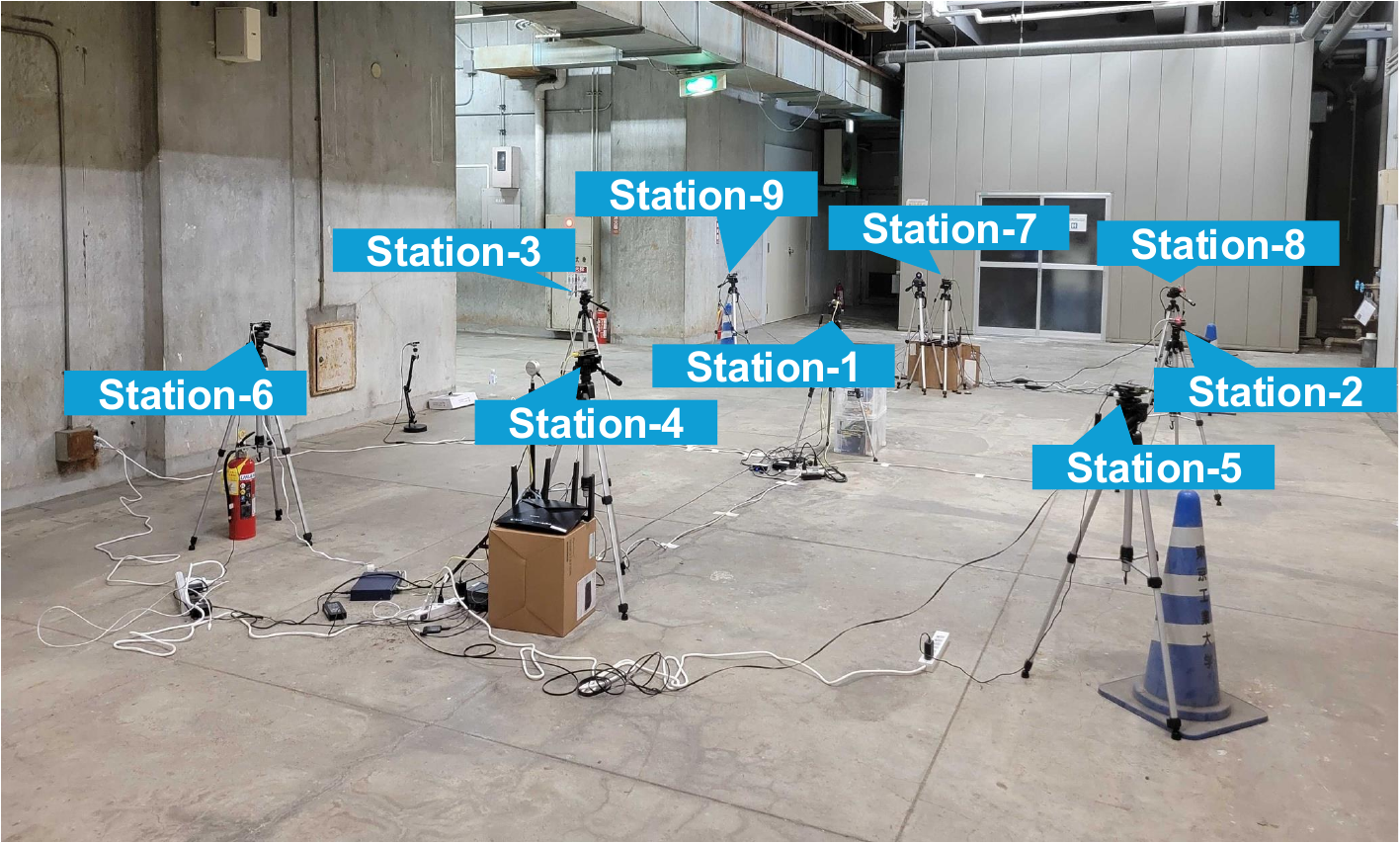}
    \caption{Snapshot 1.}
  \end{subfigure}
  \begin{subfigure}[b]{0.32\linewidth}
    \centering
    \includegraphics[width=\linewidth]{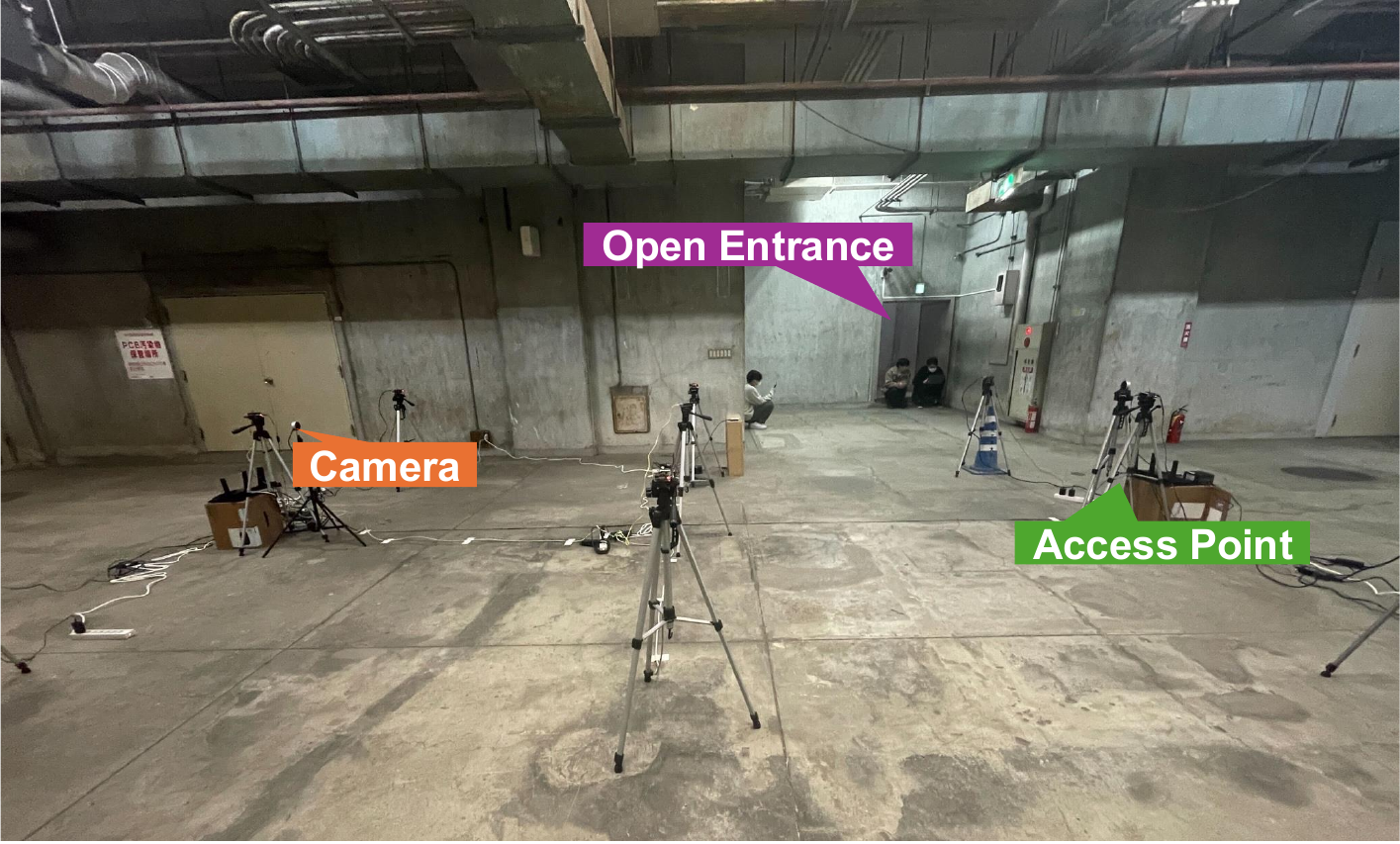}
    \caption{Snapshot 2.}
  \end{subfigure}
  \begin{subfigure}[b]{0.32\linewidth}
    \centering
    \includegraphics[width=\linewidth]{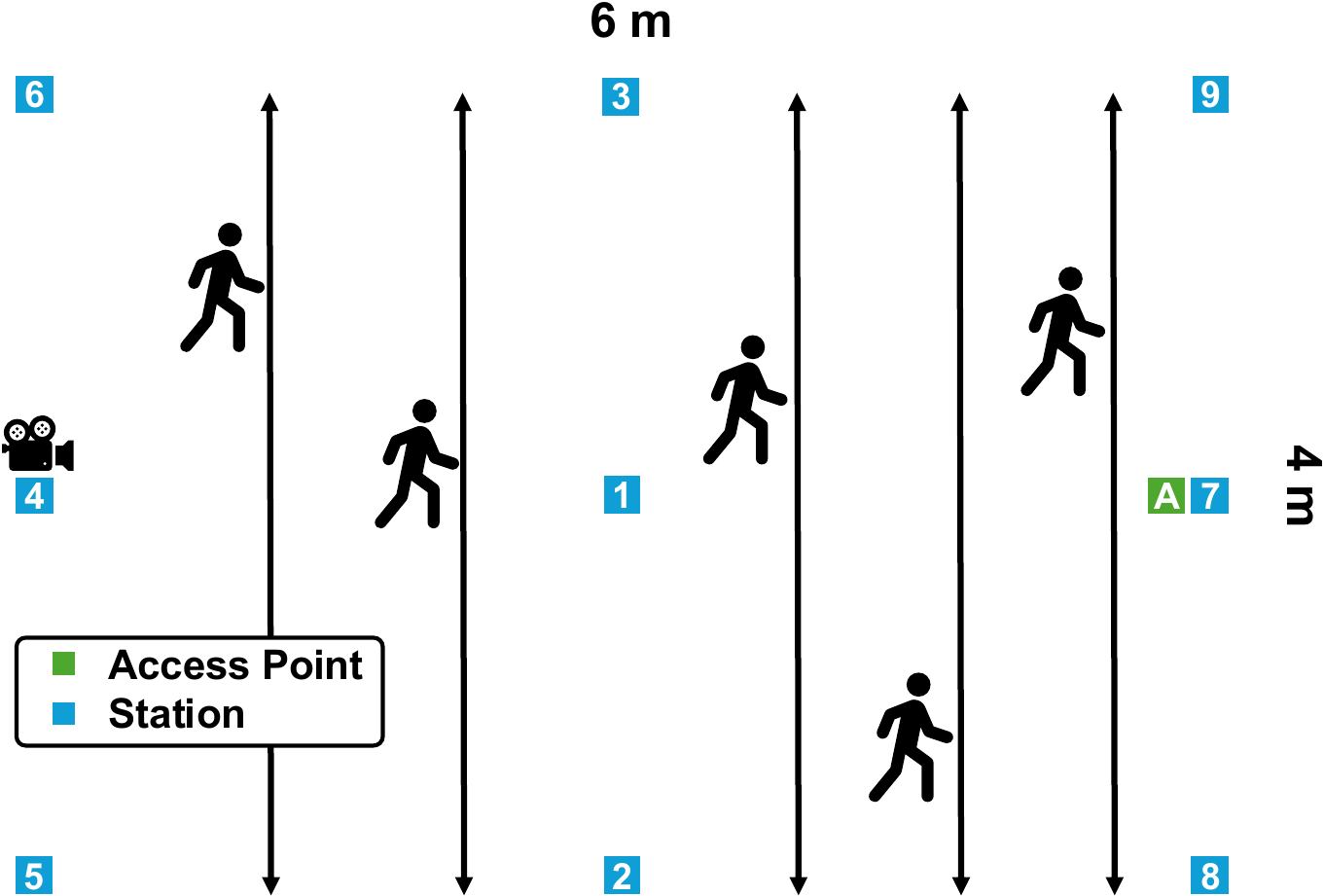}
    \caption{Layout}
  \end{subfigure}
  \caption{Experimental setup of factory-like environment.}
  \label{fig:factory_setup}
\end{figure*}

\subsection{Setup}
\noindent\textbf{Experimental setup for multi-station CSI data acquisition:}
The experimental setup and layout are illustrated in Fig.~\ref{fig:factory_setup}.
The experiment was conducted in a $4\,\mathrm{m} \times 6\,\mathrm{m}$ area.
A transmitter, acting as an AP, was placed at one end of the area.
The AP sent burst signals compliant with the IEEE 802.11ac standard in the 5\,GHz band, while five pedestrians walked along the paths indicated by black arrows.
Nine CSI measurement devices, referred to as Stations, were evenly placed around the experimental area.
Each station passively sniffed the frames transmitted by the AP and measured CSI using Nexmon CSI\cite{nexmon:paper, nexmon:project}.
During a 20\,minute experiment, single-antenna CSI with a size of $1 \times 1 \times 256$ was collected at approximately $R_{\mathrm{CSI}} \approx 500\,\mathrm{Hz}$.
Ground-truth images were obtained using an Intel RealSense L515 camera, which captured RGB images at $R_{\mathrm{Label}} \approx 5\,\mathrm{Hz}$.
All CSI acquisition devices and the camera were synchronized via a local NTP server to ensure accurate timestamp alignment, and the synchronization error was within a few milliseconds.
Table~\ref{tab:dataset} summarizes the settings used in this experiment.

Although CSI acquisition was performed by passively sniffing frames transmitted from the AP to the stations, this configuration effectively emulates the station-to-AP uplink communication scenario assumed in our system model.
This is because the wireless channel between the AP and each station can be assumed to be symmetrical, making the measured CSI largely independent of the transmission direction.

In the assumed deployment scenario, asynchronous transmissions and device unavailability can cause station-wise feature missingness at inference time.
However, in the experimental setup, all stations receive frames simultaneously from the AP, and no such missingness naturally occurs during data collection.
To ensure consistency with the assumed scenario and to rigorously evaluate robustness, station-wise feature masking was explicitly applied during training and inference.
This enables a fair assessment of the proposed method under realistic missingness conditions, despite the synchronized data acquisition.

\vspace{1mm}
\noindent\textbf{Downstream task:}
An image generation dataset was constructed from the collected CSI and image data. 
For label data, we cropped the region of interest from the RGB images, resized it to $64 \times 128 \times 3$ while preserving the aspect ratio, and normalized the pixel values to [0, 1].
The collected CSI was preprocessed following same procedure as office-like environment dataset.
As a result, the CSI was represented as $208$-dimensional real-valued vector.

\vspace{1mm}
\noindent\textbf{Dataset construction:}
The 20-minute dataset was split into training, validation, and test sets with a ratio of 7:1.5:1.5 along the time axis.
Unlabeled samples were generated at $R_{\mathrm{SSL}} = 500\,{\mathrm{Hz}}$ with a temporal window $w = 0.1\,\mathrm{s}$, producing 420,000\,samples in total, all of which contained CSI from all nine stations.
Labeled samples were generated at $R_{\mathrm{Label}} = 5\,\mathrm{Hz}$ with the same temporal window.
The resulting dataset consisted of 4,200, 900, and 900 labeled samples for training, validation, and testing, respectively, with no station missingness in any sample.

\vspace{1mm}
\noindent\textbf{Model configuration and training details:}
The constructed dataset was then used to train and evaluate machine learning models. 
For the factory-like environment dataset, both the station-wise encoder and the aggregator were implemented using three repeated blocks, each consisting of a dense layer followed by a ReLU activation, batch normalization, and dropout.
The downstream model was a CNN composed of upsampling and convolutional layers.
The size of the feature extractor and downstream model were approximately 29\,MB and 1.8\,MB when stored in single-precision floating-point format.
Both pre-training and downstream model training were performed with a batch size of 512.
As a loss function of downstream model training, we used mean squared error.
During downstream model training, unlike the office-like environment dataset, the weights of the feature extractor were updated jointly with the downstream model, whereas the other training configurations were kept identical to those used for the office-like environment dataset.

\subsection{Baseline}
The same baseline methods described in SubSection~\ref{subsec:baseline_office} were used for the factory-like environment dataset, with three modifications.
First, in the Constant baseline, the model outputs a fixed background image without any people as the predicted image.
Second, the fully supervised baselines employ a randomly initialized feature extractor, using the same architecture as DAE baseline, which is trained end-to-end with the downstream model.
Third, for data augmentation, we adopted online augmentation, where each training sample is augmented with a probability of $p_{\mathrm{aug}} = 0.5$.


\begin{figure*}[]
  \centering



  \begin{subfigure}[b]{0.24\linewidth}
    \centering
    \includegraphics[width=\linewidth]{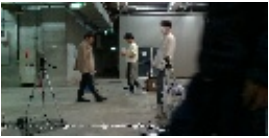}
    \caption{Ground truth}
    \label{subfig:sample_wo_label_factory}
  \end{subfigure}
  \hfill
  \begin{subfigure}[b]{0.24\linewidth}
    \centering
    \includegraphics[width=\linewidth]{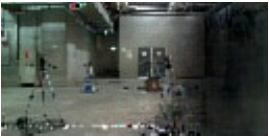}
    \caption{1 station ($\mathrm{RMSE}=0.1116$)}
  \end{subfigure}
  \hfill
  \begin{subfigure}[b]{0.24\linewidth}
    \centering
    \includegraphics[width=\linewidth]{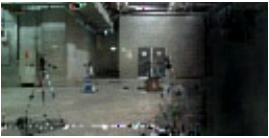}
    \caption{4 stations ($\mathrm{RMSE}=0.1049$)}
    \label{subfig:sample_wo_label_factory}
  \end{subfigure}
  \hfill
  \begin{subfigure}[b]{0.24\linewidth}
    \centering
    \includegraphics[width=\linewidth]{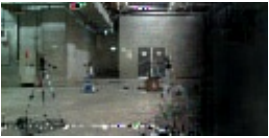}
    \caption{9 stations ($\mathrm{RMSE}=0.1057$)}
  \end{subfigure}

  \vspace{2mm}

  \begin{subfigure}[b]{0.24\linewidth}
    \centering
    \includegraphics[width=\linewidth]{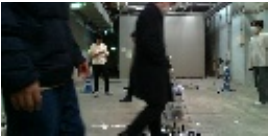}
    \caption{Ground truth}
    \label{subfig:sample_wo_label_factory}
  \end{subfigure}
  \hfill
  \begin{subfigure}[b]{0.24\linewidth}
    \centering
    \includegraphics[width=\linewidth]{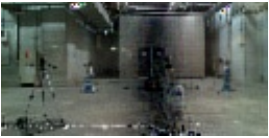}
    \caption{1 station ($\mathrm{RMSE}=0.1998$)}
  \end{subfigure}
  \hfill
  \begin{subfigure}[b]{0.24\linewidth}
    \centering
    \includegraphics[width=\linewidth]{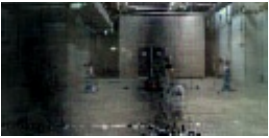}
    \caption{4 stations ($\mathrm{RMSE}=0.1255$)}
    \label{subfig:sample_wo_label_factory}
  \end{subfigure}
  \hfill
  \begin{subfigure}[b]{0.24\linewidth}
    \centering
    \includegraphics[width=\linewidth]{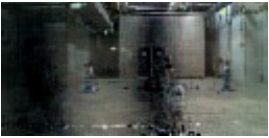}
    \caption{9 stations ($\mathrm{RMSE}=0.1257$)}
  \end{subfigure}

  \caption{
    Ground-truth and predicted images obtained by the proposed method using all labeled data for training, under different numbers of available stations for inference.
  }
  \label{fig:at_a_glance_factory}
\end{figure*}

\subsection{At a Glance}
Fig.~\ref{fig:at_a_glance_factory} provides a qualitative view of the proposed method under different levels of station-wise feature missingness.
Station-1 is used for the 1-station case, Stations-1, 2, 3 and 4 are used for the 4-stations case, and all stations are used for the 9-stations case.
Even when the number of available station is reduced, the proposed method continues to generate reasonable images without a substantial loss of overall performance.
This qualitative trend supports the robustness of the proposed method to station-wise feature missingness.

Between the two samples, the two subjects in the foreground are reconstructed relatively well, whereas the three subjects in the background are less clearly reproduced.
The limitation is likely attributable to the use of a simple CNN-based downstream model.
Because the background subjects occupy a smaller portion of the image and their clothing colors are closer to the background, their contribution to the reconstruction loss becomes relatively small. As a result, the downstream model tends to prioritize the dominant foreground regions during training, leading to weaker reconstruction of the background subjects.
A detailed investigation of image quality is beyond the scope of this study, and our focus here is on demonstrating stable inference under station-wise feature missingness.


\begin{figure}
    \centering
    \includegraphics[width=\linewidth]{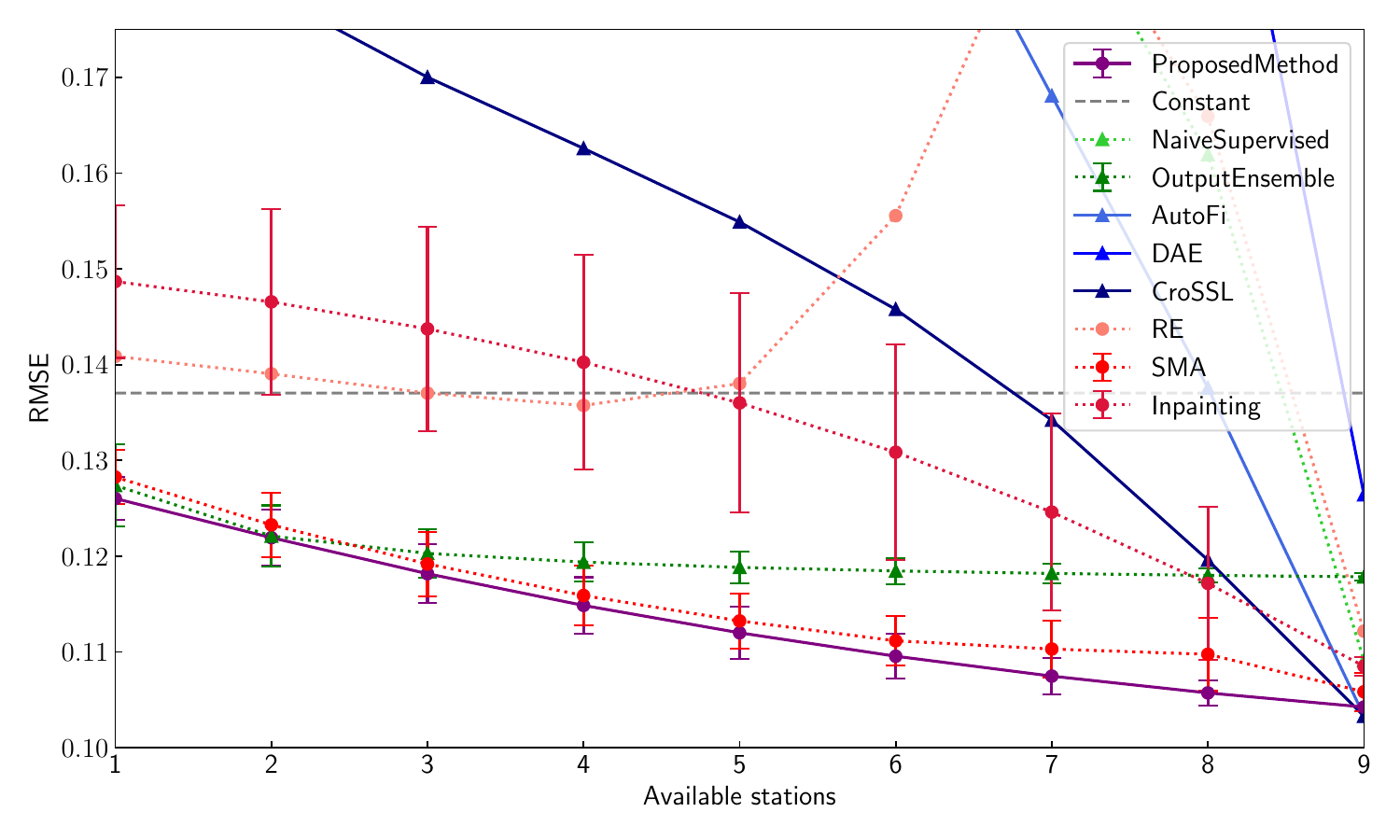}
    \caption{
    Performance comparison under different numbers of available stations on the factory-like environment dataset.
    The proposed method is shown as the purple solid line.
    For methods whose worst-case RMSE exceeds 0.175, the standard deviation bars are omitted for clarity.
    The proposed method maintains consistently lower RMSE than the baselines as stations become unavailable, demonstrating robustness to station-wise feature missingness in the factory-like environment.
    }
    \label{fig:evalu_missing_device_factory}
\end{figure}
\subsection{Robustness to Station-wise Feature Missingness}

Fig.~\ref{fig:evalu_missing_device_factory} shows the performance under different numbers of available stations.
Across almost all missingness levels, the proposed method achieves the lowest RMSE, indicating strong robustness to station-wise feature missingness.

Unlike the office-like environment, a clear performance gap emerges among missingness-aware methods in the factory-like environment.
While CroSSL alone exhibits noticeable performance degradation as the number of available stations decreases, the proposed method and SMA maintain stable performance even under severe station-wise feature missingness.
This difference suggests that learning missingness-invariant representations during pre-training alone is insufficient in more complex downstream task.
In such scenarios, explicitly exposing the downstream model to station-wise missingness becomes critical.
By jointly combining missingness-aware pre-training with station-wise masking augmentation, the proposed method effectively bridges this gap, enabling robust inference even when a large fraction of stations are unavailable.

\begin{figure}
    \centering
    \includegraphics[width=\linewidth]{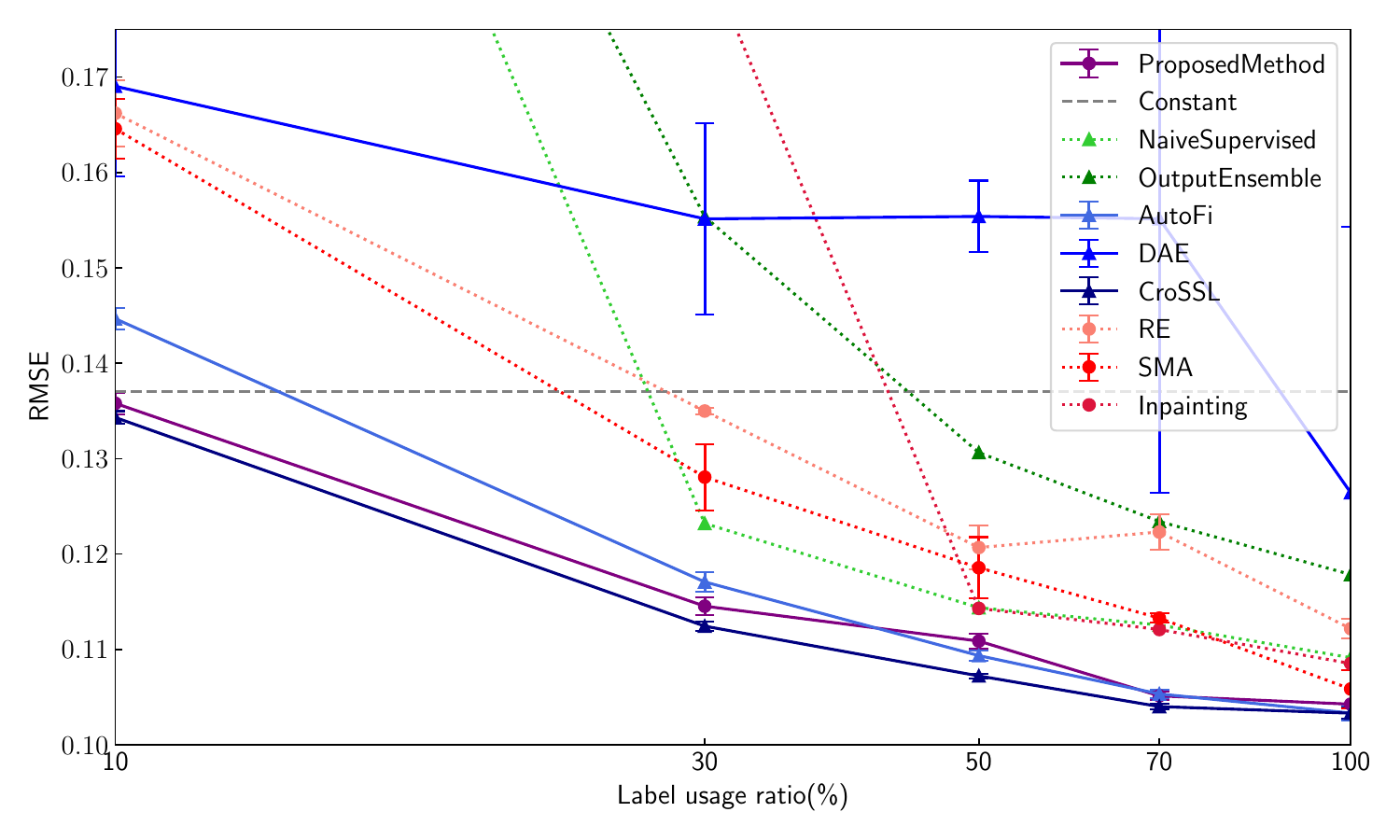}
    \caption{Performance comparison under different amounts of labeled data used for downstream model training on the factory-like environment dataset.
    For methods whose worst-case RMSE exceeds 0.175, the standard deviation bars are omitted for clarity.
     While the performance of most baselines degrades as the amount of labeled data decreases, the proposed method maintains stable performance and remains competitive across all label usage ratios, demonstrating robustness to limited labeled data.}
    \label{fig:eval_limited_label_factory}
\end{figure}
\subsection{Robustness to Limited Labeled Data}

Fig.~\ref{fig:eval_limited_label_factory} presents the performance under varying amounts of labeled data for downstream model training.
As the label usage ratio decreases, the proposed method, CroSSL, and AutoFi exhibit small performance degradation.
In contrast, fully supervised and augmentation-based baselines degrade significantly as the amount of labeled data decreases.

These results suggest that effective pre-training plays a crucial role in enabling robust learning under label scarcity, and that this benefit becomes even more important for challenging downstream tasks.

\begin{table}[]
\centering
\caption{RMSE comparison on the factory-like environment dataset, evaluating combined robustness to label scarcity and station-wise feature missingness. Best results for each condition are in bold.
For all conditions, proposed method show the best performance.}

\begin{tabular}{@{} l *{6}{c} @{}}
    \toprule
    Label usage ratio(\%)
    & \multicolumn{2}{c}{30}
    & \multicolumn{2}{c}{50} \\
    \cmidrule(lr){2-3}\cmidrule(lr){4-5}
    Available stations& 1 & 4 & 1 & 4 \\
    \midrule
    ProposedMethod & $\mathbf{0.1310}$ & $\mathbf{0.1235}$ & $\mathbf{0.1287}$ & $\mathbf{0.1202}$
    \\
    \cdashline{1-5}
    Constant & $0.1370$ & $0.1370$ & $0.1370$ & $0.1370$ 
    \\ 
    NaiveSupervised & $0.2522$ & $0.2309$ & $0.2596$ & $0.2340$
    \\
    OutputEnsemble  & $0.1746$ & $0.1592$ & $0.1375$ & $0.1317$
    \\
    \cdashline{1-5}
    AutoFi   & $0.1973$ & $0.1793$ & $0.3438$ & $0.2364$
    \\
    DAE      & $1.4721$ & $0.978$ & $1.3113$ & $0.9130$
    \\
    CroSSL   & $0.1702$ & $0.1561$ & $0.1555$ & $0.1551$
    \\
    \cdashline{1-5}
    RE  & $0.1552$ & $0.1457$ & $0.1475$ & $0.1423$
    \\
    SMA & $0.1391$ & $0.1311$ & $0.1347$ & $0.1257$
    \\
    Inpainting & $0.2125$ & $0.2030$ & $0.1585$ & $0.1441$
    \\
    \bottomrule
\end{tabular}

\label{tab:factory_combined_performance}
\end{table}

\subsection{Combined Robustness to Station-wise Feature Missingness and Limited Labeled Data}

We further evaluate the proposed method under the combined challenges of station-wise feature missingness and limited labeled data on the factory-like environment dataset.
Table~\ref{tab:factory_combined_performance} summarizes the RMSE performance for different combinations of available stations and label usage ratios, focusing on regimes where learning-based methods are expected to produce non-trivial predictions.

Under these conditions, the proposed method consistently achieves the best performance across all evaluated scenarios.
This result demonstrates that the proposed method remains effective when both incomplete multi-station observations and label scarcity occur simultaneously.

A closer inspection reveals clear limitations of methods that address only one of the two challenges.
SMA exhibits a certain degree of robustness to station-wise feature missingness; however, its performance degrades as the amount of labeled data decreases.
This behavior is expected, as SMA relies solely on data augmentation and does not leverage unlabeled data through pre-training.
Conversely, CroSSL suffers noticeable performance degradation when station-wise feature missingness occurs in low-label regimes.

In contrast, the proposed method combines missingness-aware pre-training with station-wise masking during downstream model training, enabling the downstream model to explicitly handle missing stations while benefiting from representations learned from unlabeled data.
As a result, it achieves stable and consistently strong performance under the simultaneous presence of station-wise feature missingness and limited labeled data.
These results highlight the importance of jointly addressing both challenges, particularly in the complex downstream task.

\begin{table*}[]
\centering
\caption{
RMSE comparison of different combinations of pre-training and data augmentation strategies on the factory-like environment dataset under varying numbers of available stations and label usage ratios.
For each condition, the reported RMSE is averaged over all combinations of station-missingness patterns.
Best results for each condition are shown in bold.
The results demonstrate that the proposed method consistently achieves low error across different station availability and label usage ratios.}
\begin{tabular}{@{} l *{9}{c} @{}}
    \toprule
    Label usage ratio(\%)
    & \multicolumn{3}{c}{30}
    & \multicolumn{3}{c}{50}
    & \multicolumn{3}{c}{100} \\
    \cmidrule(lr){2-4}\cmidrule(lr){5-7}\cmidrule(lr){8-10}
    Available stations& 1 & 4 & 9 & 1 & 4 & 9 & 1 & 4 & 9
    \\
    \midrule
    ProposedMethod & $\mathbf{0.1310}$ & $0.1235$ & $\mathbf{0.1145}$ & $0.1287$ & $0.1202$ & $0.1109$ & $0.1260$ & $0.1149$ & $0.1043$
    \\
    AutoFi × SMA & $\mathbf{0.1310}$ & $\mathbf{0.1230}$ & $0.1165$ & $\mathbf{0.1279}$ & $\mathbf{0.1177}$ & $\mathbf{0.1097}$ & $\mathbf{0.1252}$ & $\mathbf{0.1125}$ & $\mathbf{0.1037}$
    \\ 
    DAE × SMA    & $0.2532$ & $0.2554$ & $0.2364$ & $0.1847$ & $0.1818$ & $0.1685$ & $0.1789$ & $0.1741$ & $0.1582$
    \\
    CroSSL × RE  & $0.1545$ & $0.1383$ & $0.1149$ & $0.1583$ & $0.1387$ & $0.1106$ & $0.1613$ & $0.1390$ & $0.1061$
    \\
    \bottomrule
\end{tabular}

\label{tab:factory_comparison_combination}
\end{table*}

\subsection{Ablation Study of Pre-training and Augmentation Strategies}

Table~\ref{tab:factory_comparison_combination} presents the results of different combinations
of pre-training and data augmentation strategies on the factory-like environment dataset.
Across all evaluated conditions, the proposed method and AutoFi × SMA consistently rank among the top-performing approaches.
Although AutoFi × SMA achieves the lowest RMSE in most settings, the proposed
method remains highly competitive across all conditions, consistently staying
close to the best result.
These results indicate that combining missingness-aware pre-training with
station-wise augmentation is effective even for the more challenging image-
generation task in the factory-like environment.

\begin{figure}
    \centering
    \includegraphics[width=\linewidth]{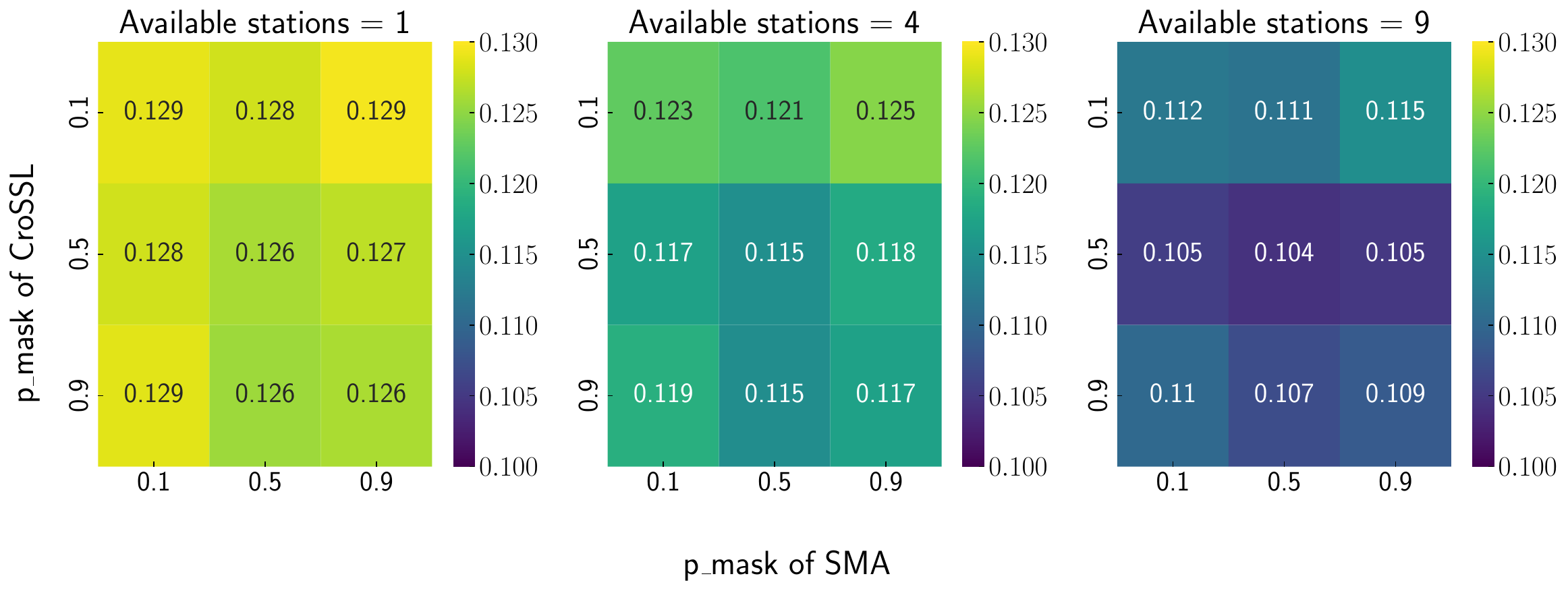}
    \caption{Each heatmap shows the RMSE for combinations of CroSSL and SMA $p_\mathrm{mask}$ values under different numbers of available stations on the factory-like environment dataset.
    Similar to the office-like environment, the combination $p_\mathrm{mask} = 0.5$ provides the most stable performance across all conditions.}
    \label{fig:masking_rate_factory}
\end{figure}
\subsection{Sensitivity Analysis of Masking Rate}

Fig.~\ref{fig:masking_rate_factory} shows a heatmap of the RMSE for various combination of CroSSL and SMA $p_{\mathrm{mask}}$.
Overall, $p_{\mathrm{mask}} = 0.5$ again provides the most stable and accurate performance across different station-availability conditions.

These results suggest that a moderate masking ratio such as $p_{\mathrm{mask}} = 0.5$ may offer an appropriate training difficulty not only in the office-like dataset but also in the factory-like dataset.
This consistency across datasets indicates that $p_{\mathrm{mask}} = 0.5$ is a promising default choice for balancing missingness simulation and representation learning.

\section{Conclusion}
In this study, we proposed a multi-station WiFi CSI sensing framework that addresses two practical challenges in real-world deployments: station-wise feature missingness
and limited labeled data.
The core idea is to learn representations that are invariant to station availability during self-supervised pre-training, and to propagate this robustness to downstream tasks through station-aware training.
Specifically, we adapted CroSSL to the multi-station CSI setting to learn
missingness-invariant representations from unlabeled data, and introduced
Station-wise Masking Augmentation (SMA) to align downstream training with realistic
station unavailability.
We evaluated the proposed framework on two real-world multi-station CSI datasets
collected in office-like and factory-like environments.
Across both datasets, the proposed method consistently outperformed conventional
supervised, augmentation-based, and existing SSL baselines over a wide range of station-missingness levels and label usage ratios.
These results indicate that explicitly incorporating station-wise missingness into both
representation learning and downstream training provides a practical and effective
foundation for deploying WiFi sensing systems in real environments, where incomplete
inputs and limited labeled data are unavoidable.

Future work includes expanding the variety of downstream tasks and model architectures to further demonstrate the generality of the proposed method. In addition, although this study assumed that signals from all stations contribute to the downstream task, this assumption may not hold in practice. Some stations may provide little useful information, such as mobile devices that continuously move. Therefore, it will be important to develop a mechanism for selecting the Stations to be used as features, including the prior exclusion of such non-contributory stations.


\bibliographystyle{IEEEtran}
\bibliography{refs}

\end{document}